\pdfoutput=1

\documentclass[10pt,twocolumn,letterpaper,sts]{article}
\usepackage{cvpr}
\usepackage{times}
\usepackage{graphicx}
\usepackage{amsmath}
\usepackage{amssymb}
\usepackage{multirow}
\usepackage{adjustbox}
 \usepackage{siunitx}
\usepackage{subcaption}
\usepackage{textcomp}
\usepackage{xcolor}
\usepackage{siunitx}
\usepackage{authblk}
\usepackage{url}
\makeatletter
\renewcommand\AB@affilsepx{ \protect\Affilfont}
\makeatother

 \makeatletter
 \newcommand\footnoteref[1]{\protected@xdef\@thefnmark{\ref{#1}}\@footnotemark}
 \makeatother

\usepackage{amsmath}
\usepackage{bm}

 \cvprfinalcopy 

\ifcvprfinal\fi
\begin{document}

\title{SfSNet: Learning Shape, Reflectance and Illuminance of Faces in the Wild}

%

\author[1]{Soumyadip Sengupta} 
\author[2]{Angjoo Kanazawa}
\author[1]{Carlos D. Castillo}
\author[1]{David W. Jacobs} 

\affil[1]{University of Maryland, College Park \hspace{2mm}} 
\affil[2]{University of California, Berkeley\\}

\maketitle

\begin{abstract}
	We present SfSNet, an end-to-end learning framework for producing an accurate decomposition of an unconstrained human face image into shape, reflectance and illuminance. SfSNet is designed to reflect a physical lambertian rendering model. SfSNet learns from a mixture of labeled synthetic and unlabeled real world images. This allows the network to capture low frequency variations from synthetic and high frequency details from real images through the photometric reconstruction loss. SfSNet consists of a new decomposition architecture with residual blocks that learns a complete separation of albedo and normal. This is used along with the original image to predict lighting. SfSNet produces significantly better quantitative and qualitative results than state-of-the-art methods for inverse rendering and independent normal and illumination estimation.
\end{abstract}

\section{Introduction}
\label{sec:intro}
In this work, we propose a method to decompose unconstrained real world faces into shape, reflectance and illuminance assuming lambertian reflectance. This decomposition or inverse rendering is a classical and fundamental problem in computer vision \cite{tappen2003recovering,prados2006shape,oxholm2012shape,barron2015shape}. It allows one to edit an image, for example with re-lighting and light transfer \cite{wang2009face}. Inverse rendering also has potential applications in Augmented Reality, where it is important to understand the illumination and reflectance of a human face. A major obstacle in solving this decomposition or any of its individual components for real images is the limited availability of ground-truth training data. Even though it is possible to collect real world facial shapes, it is extremely difficult to build a dataset of reflectance and illuminance of images in the wild at a large scale. Previous works have attempted to learn surface normal from synthetic data \cite{tran2016regressing,sela2017unrestricted}, which often performs imperfectly in the presence of real world variations like illumination and expression. Supervised learning can generalize poorly if real test data comes from a different distribution than the synthetic training data.

\begin{figure}[t]
	\centering
	\includegraphics[width=0.48\textwidth]{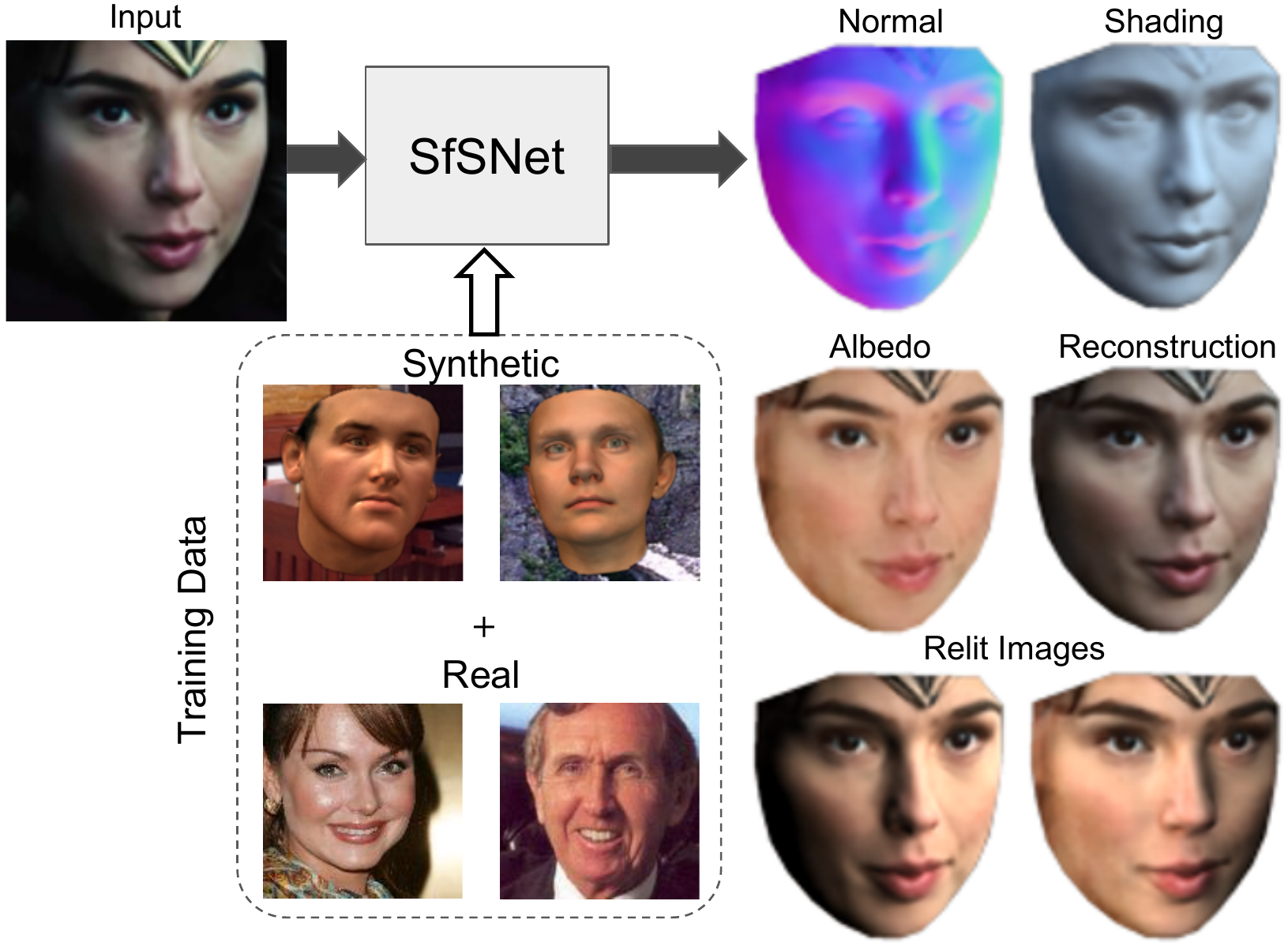}
	\caption{\small \textbf{Decomposing real world faces into shape, reflectance and illuminance.} We present SfSNet that learns from a combination of labeled synthetic and unlabeled real data to produce an accurate decomposition of an image into surface normals, albedo and lighting. Relit images are shown to highlight the accuracy of the decomposition. (Best viewed in color)}
	\vspace{-1em}
	\label{fig:teaser}	
\end{figure}

\begin{figure*}[h]
	\centering
	\includegraphics[width=0.9\textwidth]{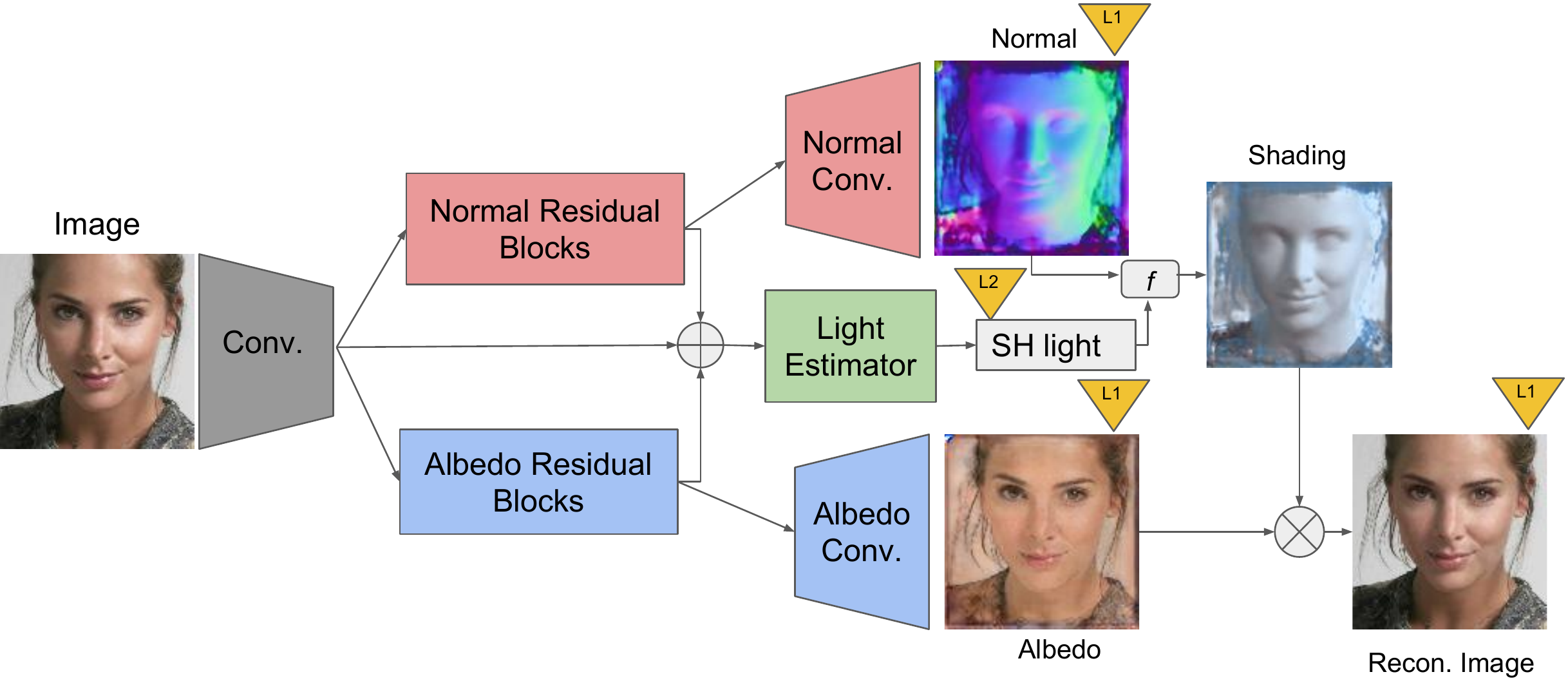}
	\caption{\small : \textbf{Network Architecture.} Our SfSNet consists of a novel decomposition architecture that uses residual blocks to produce normal and albedo features. They are further utilized along with image features to estimate lighting, inspired by a physical rendering model. $f$ combines normal and lighting to produce shading. (Best viewed in color)}
	\vspace{-1em}
	\label{fig:sfsnet}	
\end{figure*}

We propose a solution to this challenge by jointly learning all intrinsic components of the decomposition from real data. In the absence of ground-truth supervision for real data, photometric reconstruction loss can be used to validate the decomposition. This photometric consistency between the original image and inferred normal, albedo and illuminance provide strong cues for inverse rendering. However it is not possible to learn from real images only with reconstruction loss, as this may cause the individual components to collapse on each other and produce trivial solutions. Thus, a natural step forward is to get the best of both worlds by simultaneously using supervised data when available and real world data with reconstruction loss in their absence. To this end we propose a training paradigm `SfS-supervision'.

To achieve this goal we propose a novel deep architecture called SfSNet, which
attempts to mimic the physical model of lambertian image generation while
learning from a mixture of labeled synthetic and unlabeled real world
images. Training from this mixed data allows the network to learn low frequency
variations in facial geometry, reflectance and lighting from synthetic data
while simultaneously understanding the high frequency details in real data using
shading cues through reconstruction loss. This idea is motivated by the classical works in the Shape from
Shading (SfS) literature where often a reference model is used to compensate for
the low frequency variations and then shading cues are utilized for obtaining
high frequency details \cite{kemelmacher20113d}. To meet this goal we develop a
decomposition architecture with residual blocks that learns a complete
separation of image features into normals and albedo. Then we use normal, albedo and
image features to regress the illumination parameters. This is based on the
observation that in classical illumination modeling, lighting is estimated
from image, normal and albedo by solving an over-constrained system of
equations. Our network architecture is illustrated in  
Figure \ref{fig:sfsnet}.  Our model and code is available for research purposes at {\small \url{https://senguptaumd.github.io/SfSNet/}}.

We evaluate our approach on the real world CelebA dataset
\cite{liu2015faceattributes} and present extensive comparison with recent state-of-the-art methods
\cite{adobe,mofa}, which also perform inverse rendering of faces. SfSNet produces significantly better reconstruction than \cite{adobe,mofa} on the same images that are showcased in their papers. We further compare SfSNet with state-of-the-art methods that aim to solve
for only one component of the inverse rendering such as normals or lighting. SfSNet
outperforms a recent approach that estimates normal independently \cite{trigeorgis2017normals}, by improving normal estimation accuracy
by 47\% (37\% to 84\%) on the Photoface dataset \cite{zafeiriou2011photoface}, which contains faces captured under harsh lighting. We also compare against ‘Pix2Vertex’ \cite{sela2017unrestricted}, which only estimate high resolution meshes. We demonstrate that SfSNet reconstructions are significantly more robust to expression and illumination variation compared to ‘Pix2Vertex’. This results from the fact that we are jointly solving for all components, which allows us to train on real images through reconstruction loss. SfSNet outperforms `Pix2Vertex' (before meshing) by 19\% (25\% to 44\%) without training on the Photoface dataset. We also outperform a recent approach on lighting estimation `LDAN' \cite{zhou2017label} by 12.5\% (65.9\% to 78.4\%).

In summary our main contributions are as follows:\\
$(1)$ We propose a network, SfSNet, inspired by a physical lambertian rendering model. This uses a decomposition architecture with residual blocks to separate image features into normal and albedo, further used to estimate lighting.\\
$(2)$ We present a training paradigm `SfS-supervision', which allows learning from a mixture of labeled synthetic and unlabeled real world images. This allows us to jointly learn normal, albedo and lighting from real images via reconstruction loss, outperforming approaches that only learn an individual component.\\
$(3)$ SfSNet produces remarkably better visual results compared to state-of-the-art methods for inverse rendering \cite{adobe,mofa}. In comparison with methods that obtain one component of the inverse rendering \cite{trigeorgis2017normals,sela2017unrestricted,zhou2017label}, SfSNet is significantly better, especially for images with expression and non-ambient illumination.

\section{Related Work}
\label{sec:related}
\textbf{Classical approaches for inverse rendering:} The problem of decomposing
shape, reflectance and illuminance from a single image is a classical problem in
computer vision and has been studied in various forms such as intrinsic image
decomposition \cite{tappen2003recovering} and Shape from Shading (SfS)
\cite{prados2006shape,oxholm2012shape}. Recent work of SIRFS
\cite{barron2015shape} performs decomposition of an object into surface normal,
albedo and lighting assuming lambertian reflection by formulating extensive
priors in an optimization framework. The problem of inverse rendering in the
form of SfS gained particular attention in the domain of human facial
modeling. This research was precipitated by the advent of the 3D Morphable Model
(3DMM) \cite{blanz1999morphable} as a potential prior for shape and
reflectance. Recent works used facial priors to reconstruct shape from a single
image
\cite{kemelmacher2011face,kemelmacher20113d,chai2015high,saito2016realtime} or
from multiple images \cite{roth2016adaptive}. Classical SfS methods fail to
produce realistic decomposition on unconstrained real images. More
recently, Saito \etal proposes a method to synthesize a photorealistic albedo
from a partial albedo obtained by traditional methods \cite{saito2016photorealistic}.

\textbf{Learning based approaches for inverse rendering:} In recent years, researchers have focused on data driven approaches for learning priors rather than hand-designing them for the purpose of inverse rendering. Attempts at learning such priors were presented in \cite{tang2012deep} using Deep Belief Nets and in \cite{kulkarni2015deep} using a convolutional encoder-decoder based network. However these early works were limited in their performance on real world unconstrained faces. Recent work from Shu \etal \cite{adobe} aims to find a meaningful latent space for normals, albedo and lighting to facilitate various editing of faces. Tewari \etal \cite{mofa} solves this facial disentanglement problem by fitting a 3DMM for shape and reflectance and regressing illumination coefficients. Both \cite{adobe,mofa} learn directly from real world faces by using convolutional encoder-decoder based architectures. Decompositions produced by \cite{adobe} are often not realistic; and \cite{mofa} only captures low frequency variations.  In contrast, our method learns from a mixture of labeled synthetic and unlabeled real world faces using a novel decomposition architecture. Although our work concentrates on decomposing faces, the problem of inverse rendering for generic objects in a learning based framework has also gained attention in recent years \cite{bell2014intrinsic,narihira2015direct,shi2016learning,janner2017self}.

\textbf{Learning based approaches for estimating individual components:} Another direction of research is to estimate shape or illumination of a face independently. Recently many research works aim to reconstruct the shape of real world faces by learning from synthetic data; by fitting a 3DMM \cite{tran2016regressing,laine2017production,thies2016face}, by predicting a depth map and subsequent non-rigid deformation to obtain a mesh \cite{sela2017unrestricted} and by regressing a normal map \cite{trigeorgis2017normals}. Similarly \cite{zhou2017label} proposed a method to estimate lighting directly from a face. These learning based independent component estimation methods can not be trained with unlabeled real world data and thus suffer from the ability to handle unseen face modalities. In contrast our joint estimation approach performs the complete decomposition while allowing us to train on unlabeled real world images using our `SfS-supervision'. 

\textbf{Architectures for learning based inverse rendering:}  In \cite{adobe}, a convolutional auto-encoder was used for disentanglement and generating normal and albedo images. However recent advances in skip-connection based convolutional encoder-decoder architectures for image to image translations \cite{ronneberger2015u,isola2016image,CycleGAN2017} have also motivated their use in \cite{shi2016learning}. Even though skip connection based architectures are successful in transferring high frequency informations from input to output, they fail to produce meaningful disentanglement of both low and high frequencies. Our proposed decomposition architecture uses residual block based connections that allow the flow of high frequency information from input to output while each layer learns both high and low frequency features. A residual block based architecture was used for image to image translation in \cite{Johnson2016Perceptual} for style transfer and in a completely different domain to learn a latent subspace with Generative Adversarial Networks  \cite{liu2017unsupervised}.

\section{Approach}
Our goal is to use synthetic data with ground-truth supervision over normal, albedo and lighting along with real images with no ground-truth. We assume image formation under lambertian reflectance. Let $N(p)$, $A(p)$ and $I(p)$ denote the normal, albedo and image intensity at each pixel $p$. We represent lighting $L$ as nine dimensional second order spherical harmonics coefficients for each of the RGB channels. The image formation process under lambertian reflectance, following \cite{basri2003lambertian} is represented in equation (\ref{eq:ren1}), where $f_{\textit{render}}(.)$ is a differentiable function.
\begin{equation}
\label{eq:ren1}
I(p) = f_{\textit{render}}(N(p),A(p),L)
\end{equation}

\subsection{`SfS-supervision' Training}
\label{sec:loss}
Our `SfS-supervision' consists of a multi-stage training as follows: (a) We train a simple skip-connection based encoder-decoder network on labeled synthetic data. (b) We apply this network on real data to obtain normal, albedo and lighting estimates. These elements will be used in the next stage as `pseudo-supervision'. (c) We train our SfSNet with a mini-batch of synthetic data with ground-truth labels and real data with `pseudo-supervision' labels. Along with supervision loss over normal, albedo and lighting we use a photometric reconstruction loss that aims to minimize the error between the original image and the reconstructed image following equation \eqref{eq:ren1}.

This reconstruction loss plays a key role in learning from real data using shading cues while `pseudo-supervision' prevents the collapse of individual components of the decomposition that produce trivial solutions. In Section \ref{sec:mix_res} we show that `SfS-supervision' significantly improves inverse rendering over training on synthetic data only. Our idea of `SfS-supervision' is motivated by the classical methods in SfS, where a 3DMM or a reference shape is first fitted and then used as a prior to recover the details
\cite{kemelmacher20113d,kemelmacher2011face}.  Similarly in `SfS-supervision', low frequency variations are obtained by learning from synthetic data. Then they are used as priors or `pseudo-supervision' along with photometric reconstruction loss to add high frequency details.

Our loss function is described in equation \eqref{eq:loss}. For $E_N$, $E_A$ and $E_{recon}$ we use $L_{1}$ loss over all pixels of the face for normal, albedo and reconstruction respectively; $E_L$ is defined as the $L_2$ loss over 27 dimensional spherical harmonic coefficients. We train with a mixture of synthetic and real data in every mini-batch. We use $\lambda_{recon}$, $\lambda_N$ and $\lambda_A$ $=$ 0.5 and $\lambda_L$ $=$ 0.1. Details of reconstruction loss under lambertian reflectance are presented in the Appendix \ref{sec:sph}.
\begin{equation}
\label{eq:loss}
E = \lambda_{recon} E_{recon} + \lambda_{N} E_{N} + \lambda_A E_A +\lambda_L E_L
\end{equation}


\subsection{Proposed Architecture}
\label{sec:arch}
A common architecture in image to image translation is skip-connection based encoder-decoder networks \cite{ronneberger2015u,isola2016image}. In the context of inverse rendering, \cite{shi2016learning} used a similar skip-connection based network to perform decomposition for synthetic images consisting of ShapeNet \cite{chang2015shapenet} objects. We observe that in these networks most of the high frequency variations are passed from encoder to decoders through the skip connections. Thus the networks do not have to necessarily reason about whether high frequency variations like wrinkles and beards come from normal or albedo. Also in these networks the illumination is estimated only from the image features directly and is connected to normal and albedo through reconstruction loss only. However since illumination can be estimated from image, normal and albedo by solving an over-constrained system of equations, it makes more sense to predict lighting from image, normal and albedo features.

\begin{figure}[]
	\centering
	\includegraphics[width=0.48\textwidth]{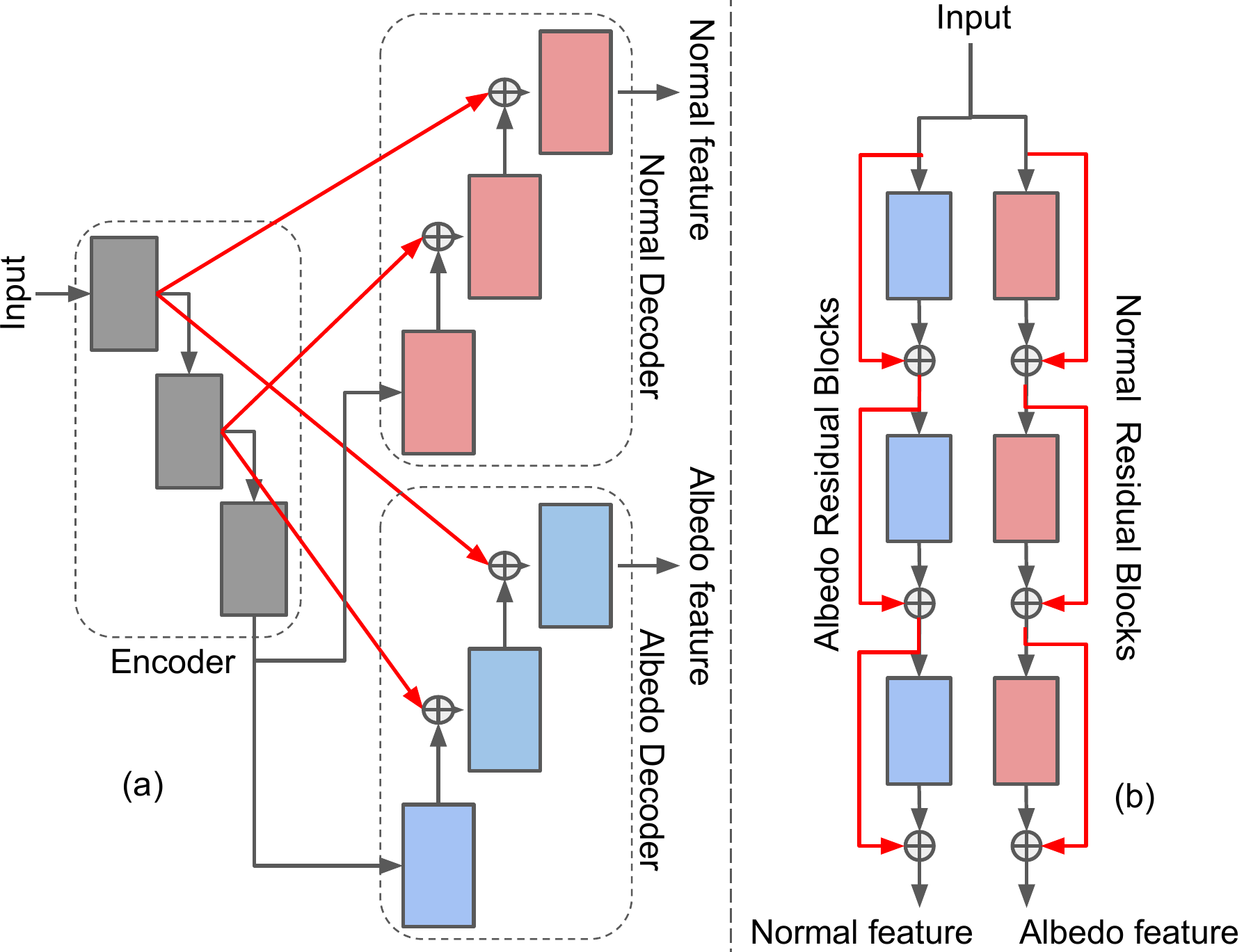}
	\caption{\textbf{Decomposition architectures.} We experiment with two architectures:   (a) skip connection based encoder-decoder; (b) proposed residual block based network. Skip connections are shown in red.}
	\vspace{-1em}
	\label{fig:skip_res}
\end{figure}

The above observations motivate us to develop an architecture that learns to separate both low and high frequency variations into normal and albedo to obtain a meaningful subspace that can be further used along with image features to predict lighting. Thus we use a residual block based architecture as shown in Figure \ref{fig:sfsnet}. The decomposition with `Normal Residual Blocks' and `Albedo Residual Blocks' allows complete separation of image features into albedo and normal features as shown in Figure \ref{fig:skip_res}b. The skip connections (shown in red) allow the high frequency information to flow directly from input feature to output feature while the individual layers can also learn from the high frequency information present in the skip connections. This lets the network learn from both high and low frequency information and produce a meaningful separation of features at the output. In contrast a skip connection based convolutional encoder-decoder network as shown in Figure \ref{fig:skip_res}a consists of skip connections (shown in red) that bypass all the intermediate layers and flow directly to the output. This architecture allows us to estimate lighting from a combination of image, normal and albedo features. In Section \ref{sec:resVsskip} we show that using a residual block based decomposition improves lighting estimation by 11\% (67.7\% to 78.4\%) compared to a skip connection based encoder-decoder.

The network uses few layers of convolution to obtain image features, denoted by $I_{f}$ which is the output of the `Conv' block in Figure \ref{fig:sfsnet}. $I_f$ is the input to two different residual blocks denoted as `Normal Residual Blocks' and `Albedo Residual Blocks', which take the image features and learns to separate them into normal and albedo features. Let  the output of `Normal Residual Blocks' and `Albedo Residual Blocks' be $N_{f}$ and $A_{f}$ respectively. $N_{f}$ and $A_{f}$ are further processed through `Normal Conv' and `Albedo Conv' respectively to obtain normal and albedo aligned with the original face. To estimate lighting we use image ($I_f$), normal ($N_f$) and albedo ($A_f$) features in the `Light Estimator' block of Figure \ref{fig:sfsnet} to obtain 27 dimensional spherical harmonic coefficients of lighting. The `Light Estimator' block simply concatenates image, normal and albedo features followed by 1x1 convolutions, average pooling and a fully connected layer to produce lighting coefficients. The details of the network are provided in the Appendix \ref{sec:arch_sfs}.

\subsection{Implementation Details}
\label{sec:imple}
To generate synthetic data we use 3DMM \cite{blanz1999morphable} in various
viewpoints, reflectance and illumination. We render these models using 27 dimensional spherical harmonics coefficients (9 for each RGB channel), which comes from a distribution estimated by fitting 3DMM  over real images from the CelebA dataset using classical methods. We use CelebA \cite{liu2015faceattributes} as real data for both training, validation and testing, following the provided protocol. For real images we detect keypoints using \cite{ranjan2017all} and create a mask based on these keypoints. Each of the ‘Residual Blocks’ consists of 5 residual blocks based on the structure proposed by \cite{he2016identity}. Our network is trained with
input images of size $128 \times 128$ and the residual blocks all operate at $64
\times 64$ resolution. The `pseudo-supervision' for real world images are generated by training a simple skip-connection based encoder-decoder network, similar to \cite{adobe}, on synthetic data. This network is also referred to as `SkipNet' in Section \ref{sec:mix_res} and details are provided in the Appendix \ref{sec:arch_skip}.


\section{Comparison with State-of-the-art Methods}
\label{sec:SOTAexp}
\begin{figure}[t]
	\centering
	\includegraphics[width=0.45\textwidth]{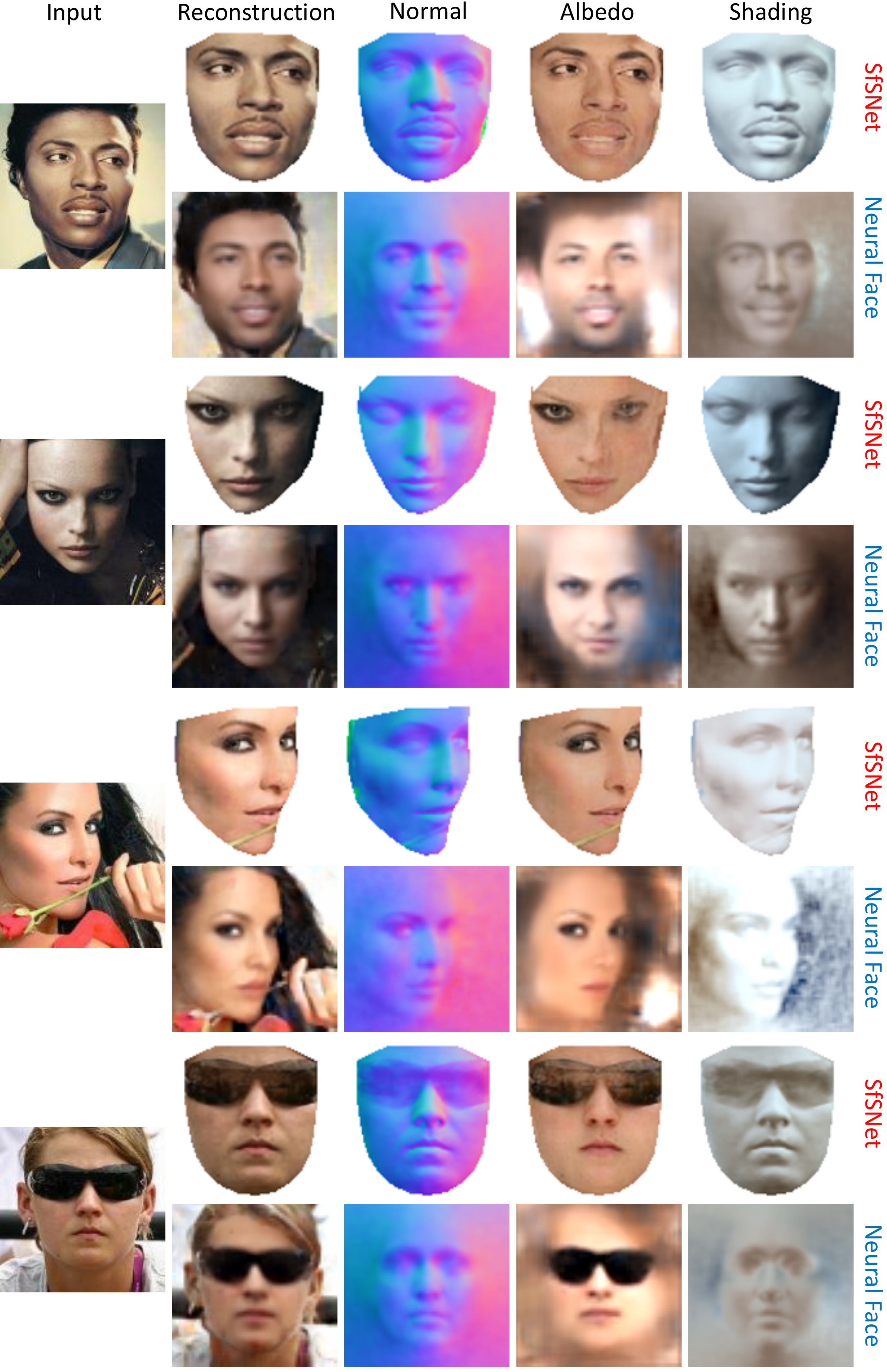}
	\caption{\small \textbf{Inverse Rendering.} \textbf{SfSNet vs `Neural Face'} \cite{adobe} on the data showcased by the authors. (Best viewed in color)}
	\vspace{-1em}
	\label{fig:adobe}	
\end{figure}
\begin{figure}[]
	\centering
	\includegraphics[width=0.48\textwidth]{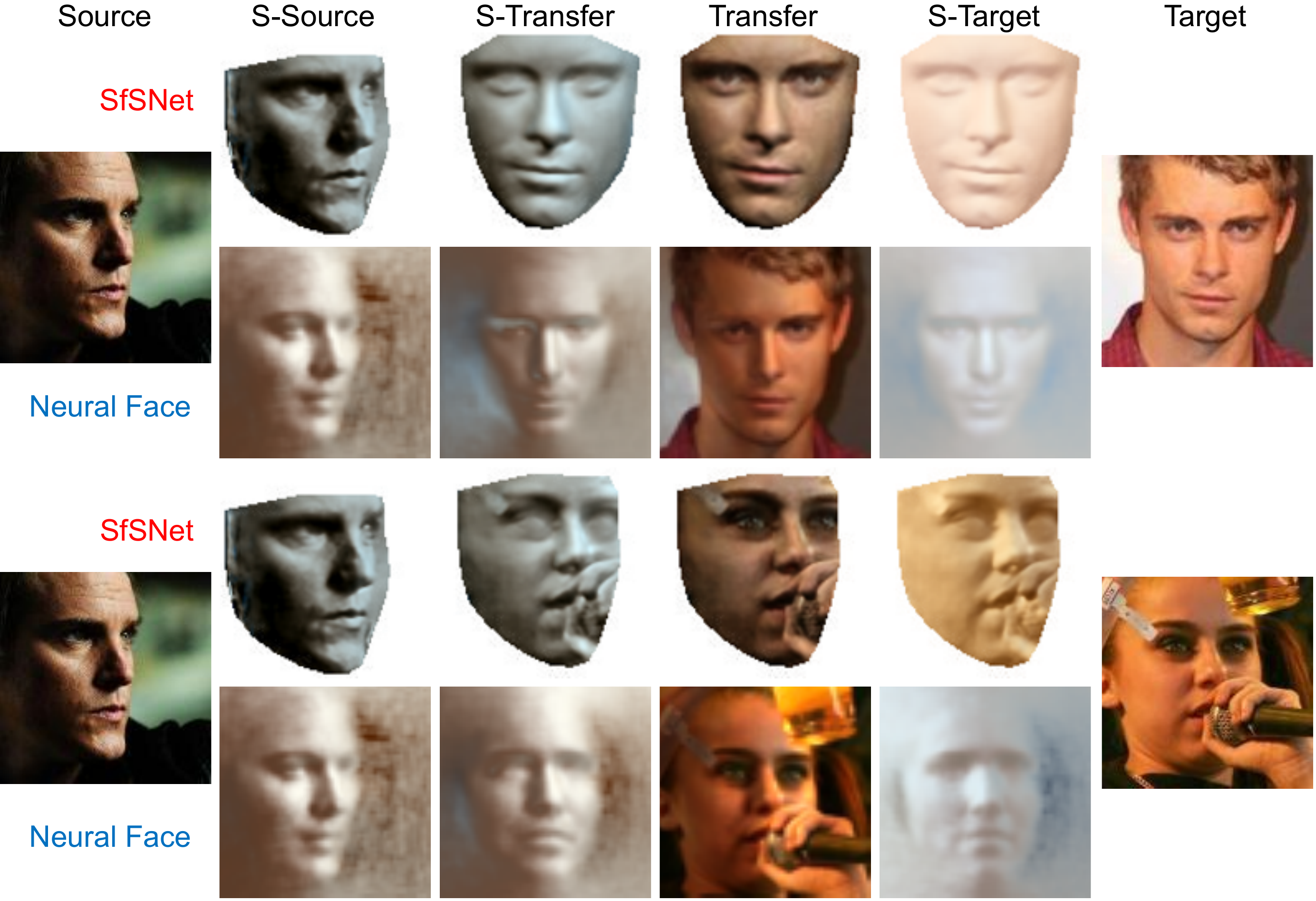}
	\caption{\small \textbf{Light Transfer.} \textbf{SfSNet vs `Neural Face'} \cite{adobe} on the image showcased by the authors. We transfer the lighting of the `Source' image to the `Target' image to produce `Transfer' image. \textit{S} denotes shading. Both `Target' images contain an orangey glow, which is not present in the `Source' image. Ideally in the `Transfer' image, the orangey glow should be removed. `Neural Face' fails to get rid of the orangey lighting effect of the `Target' image in the `Transfer' image. (Best viewed in color)}
	\vspace{-0.5em}
	\label{fig:adobe_light}	
\end{figure}
We compare our SfSNet with \cite{adobe,mofa} qualitatively on unconstrained real world faces. 
As an application of inverse rendering we perform light transfer between a pair
of images, which also illustrates the correctness of the decomposition. We  quantitatively evaluate the estimated normals on the Photoface dataset \cite{zafeiriou2011photoface} and compare with the state-of-the-art \cite{trigeorgis2017normals,sela2017unrestricted}. Similarly we also evaluate the accuracy of estimated lighting on the MultiPIE dataset \cite{gross2010multi} and compare with \cite{zhou2017label}. We outperform state-of-the-art methods by a large margin both qualitatively and quantitatively.

\subsection{Evaluation of Inverse Rendering}
\begin{figure}[]
	\centering
	\includegraphics[width=0.48\textwidth]{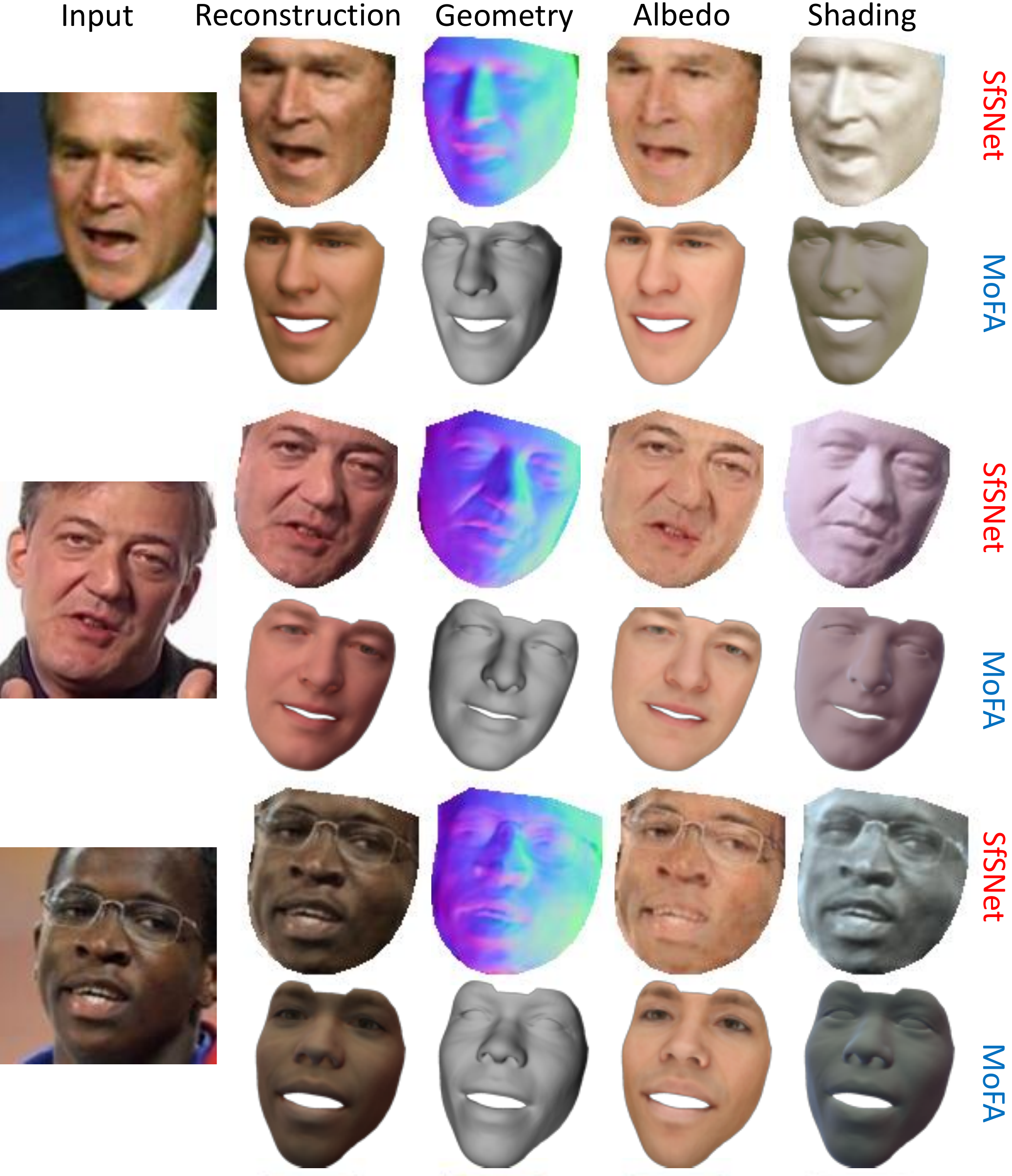}
	\caption{\small \textbf{Inverse Rendering.} \textbf{SfSNet vs `MoFA'} \cite{mofa} on the data provided by the authors of the paper. (Best viewed in color)}
	\vspace{-1em}
	\label{fig:mofa}	
\end{figure}

\begin{figure}[h]
	\centering
	\includegraphics[width=0.48\textwidth]{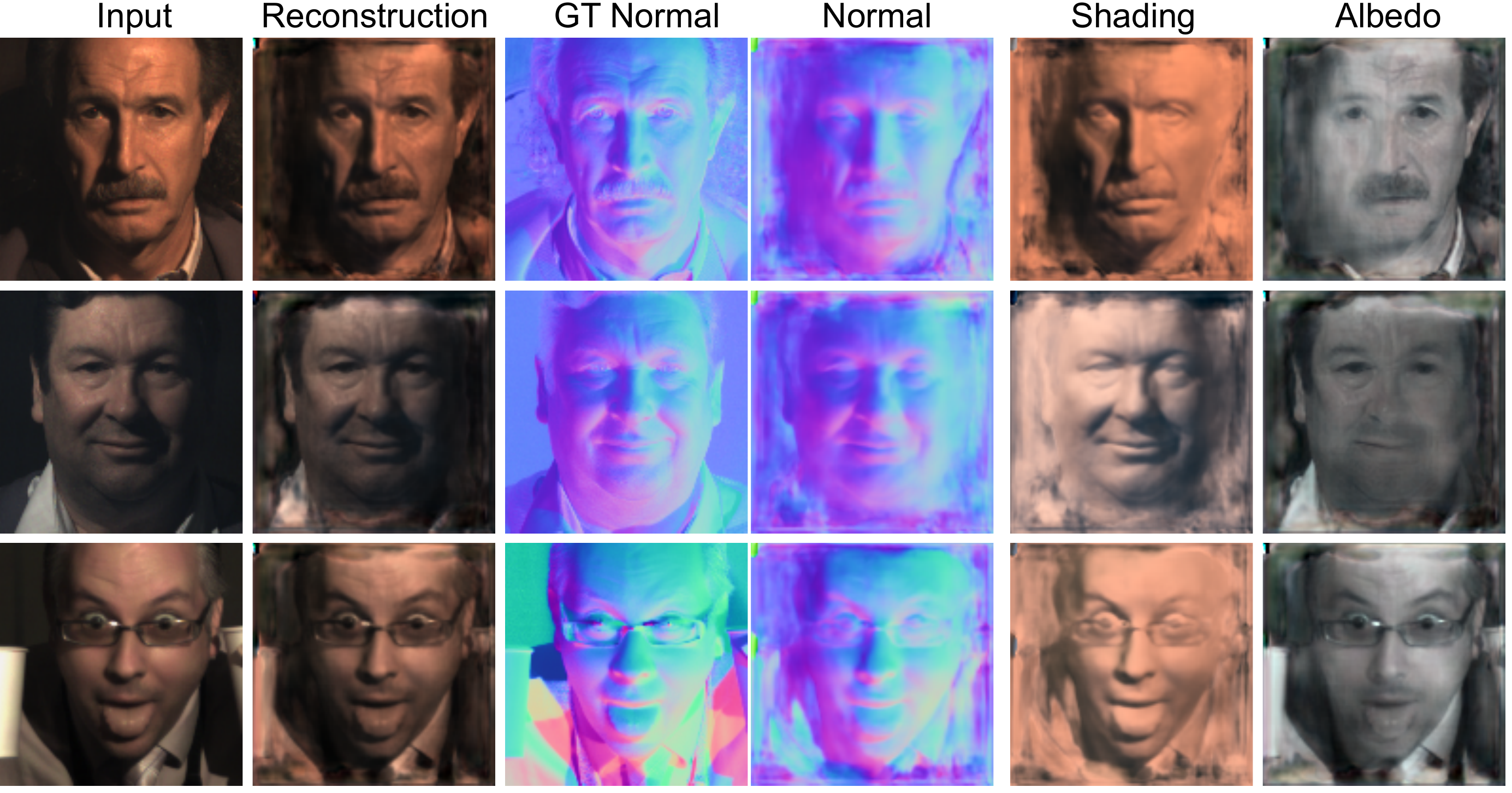}
	\caption{\small \textbf{Inverse Rendering on the Photoface dataset \cite{zafeiriou2011photoface} with `SfSNet-finetuned'.} The ground-truth albedo is in gray-scale and it encourages our network to also output gray-scale albedo.}
	\vspace{-1em}
	\label{fig:pf_fig}	
\end{figure}
\noindent
In Figures \ref{fig:adobe} and \ref{fig:adobe_light} we compare performance of our SfSNet with `Neural Face' \cite{adobe} on inverse rendering and light transfer respectively. The results are shown on the same images used in their paper. The results clearly show that SfSNet performs more realistic decomposition than `Neural Face'. Note that in light transfer `Neural Face' does not use their decomposition, but rather recomputes the albedo of the target image numerically. Light transfer results in Figure \ref{fig:adobe_light}, show that SfSNet recovers and transfers the correct ambient light compared to `Neural Face', which fails to get rid of the orangey lighting from the target images. 
We also compare inverse rendering results of SfSNet on the images provided to us by the authors of \cite{mofa} in Figure \ref{fig:mofa}. Since \cite{mofa} aims to fit a 3DMM that can only capture low frequency variations, we obtain more realistic normals, albedo and lighting than them.

\subsection{Evaluation of Facial Shape Recovery}
In this section we compare the quality of our reconstructed normals with that of current state-of-the-art methods that only recover shape from a single image. We use the Photoface dataset \cite{zafeiriou2011photoface}, which provides ground-truth normals for images taken under harsh lighting. First we compare with algorithms that also train on the Photoface dataset. We finetune our SfSNet on this dataset using ground truth normals and albedo as supervision since they are available.  We compare our `SfSNet-ft' with `NiW' \cite{trigeorgis2017normals} and other baseline algorithms, `Marr Rev.' \cite{bansal2016marr}   and `UberNet' \cite{kokkinos2016ubernet}, reported in \cite{trigeorgis2017normals} in Table \ref{tab:normal_pf}. The metric used for this task is mean angular error of the normals and the percentage of pixels at various angular error thresholds as in \cite{trigeorgis2017normals}. Since the exact training split of the dataset is not provided by the authors, we create a random split based on identity with 100 individuals in test data as mentioned in their paper. Our `SfSNet-ft' improves normal estimation accuracy by more than a factor of two for the most challenging threshold of 20 degrees accuracy. In Figure \ref{fig:pf_fig} we show visual results of decomposition on test data of the Photoface dataset.


\begin{figure}[t]
	\centering
	\includegraphics[width=0.48\textwidth]{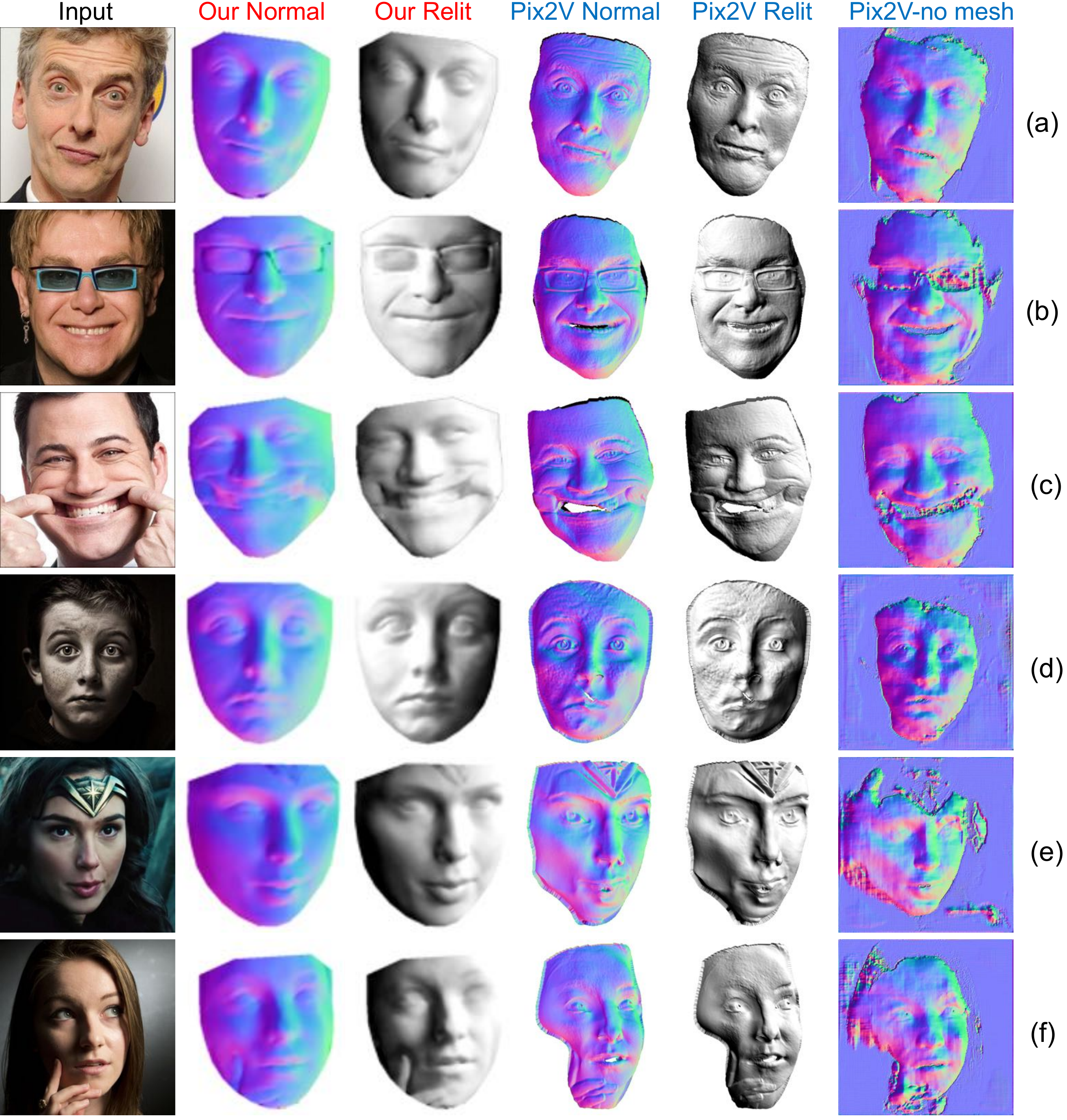}
	\caption{\small \textbf{SfSNet vs Pix2Vertex} \cite{sela2017unrestricted}. Normals produced by SfSNet are significantly better than Pix2Vertex, especially for non-ambient illumination and expression. `Relit' images are generated by directional lighting and uniform albedo to highlight the quality of the reconstructed normals. Note that (a), (b) and (c) are the images showcased by the authors. (Best viewed in color)}
	\vspace{-1em}
	\label{fig:kimmel}	
\end{figure}

\begin{table}[!h]
	\centering
	\begin{adjustbox}{max width=0.48\textwidth}
		\begin{tabular}{|c|c|c|c|c|}
			\hline
			Algorithm           & Mean \textpm std           & $<$ \ang{20}    & $<$ \ang{25}    & $<$ \ang{30}    \\ \hline
			3DMM                & 26.3 \textpm 10.2          & 4.3\%           & 56.1\%          & \textbf{89.4\%}          \\
			Pix2Vertex\cite{sela2017unrestricted} &   33.9 \textpm 5.6        &    24.8\%     &    36.1\%     &     47.6\%  \\
			SfSNet              & \textbf{25.5 \textpm 9.3}          & \textbf{43.6\%}          & \textbf{57.5\%}          & 68.7\%          \\ \hline
			Marr Rev.\cite{bansal2016marr}           & 28.3 \textpm 10.1          & 31.8\%          & 36.5\%          & 44.4\%          \\ 
			UberNet\cite{kokkinos2016ubernet}           & 29.1 \textpm 11.5          & 30.8\%          & 36.5\%          & 55.2\%          \\ 
			NiW\cite{trigeorgis2017normals} & 22.0 \textpm 6.3           & 36.6\%         & 59.8\%          & 79.6\%          \\ 
			SfSNet-ft              & \textbf{12.8 \textpm 5.4} & \textbf{83.7\%} & \textbf{90.8\%} & \textbf{94.5\%} \\ \hline
		\end{tabular}
	\end{adjustbox}
	\caption{\small \textbf{Normal reconstruction error on the Photoface dataset}. 3DMM, Pix2Vertex and SfSNet are not trained on this dataset. Marr Rev., UberNet, NiW and SfSNet-finetuned (SfSNet-ft) are trained on the training split of this dataset. Lower is better for mean error (column 1), and higher is better for the percentage of correct pixels at various thresholds (columns 3-5).}
	\vspace{-1em}
	\label{tab:normal_pf}
\end{table}

Next we compare our algorithm with `Pix2Vertex' \cite{sela2017unrestricted}, which is trained on higher resolution 512$\times$512 images. `Pix2Vertex' learns to produce a depth map and a deformation map that are post-processed to produce a mesh. In contrast our goal is to perform inverse rendering. Since we are able to train on real data, unlike `Pix2Vertex', which is trained on synthetic data, we can better capture real world variations. Figure \ref{fig:kimmel} compares normals produced by SfSNet with that of `Pix2Vertex' both before and after meshing on the images showcased by the authors. 
Since `Pix2vertex' handles larger resolution and produces meshes, their normals can capture more details than ours. But with more expression and non-ambient illumination like (c), (d), (e) and (f) in Figure \ref{fig:kimmel}, we produce fewer artifacts and more realistic normals and shading. SfSNet is around 2000$\times$ faster than `Pix2Vertex' due to the expensive mesh generation post-processing. These results show that learning all components of inverse rendering jointly allows us to train on real images to capture better variations than `Pix2Vertex'.  We further compare SfSNet with the normals produced by `Pix2Vertex' quantitatively before meshing on the Photoface dataset. SfSNet, `Pix2Vertex' and 3DMM are not trained on this dataset. The results shown in Table \ref{tab:normal_pf} shows that SfSNet outperforms `Pix2Vertex' and 3DMM by a significant margin.

\begin{figure}[]
	\centering
	\includegraphics[width=0.48\textwidth]{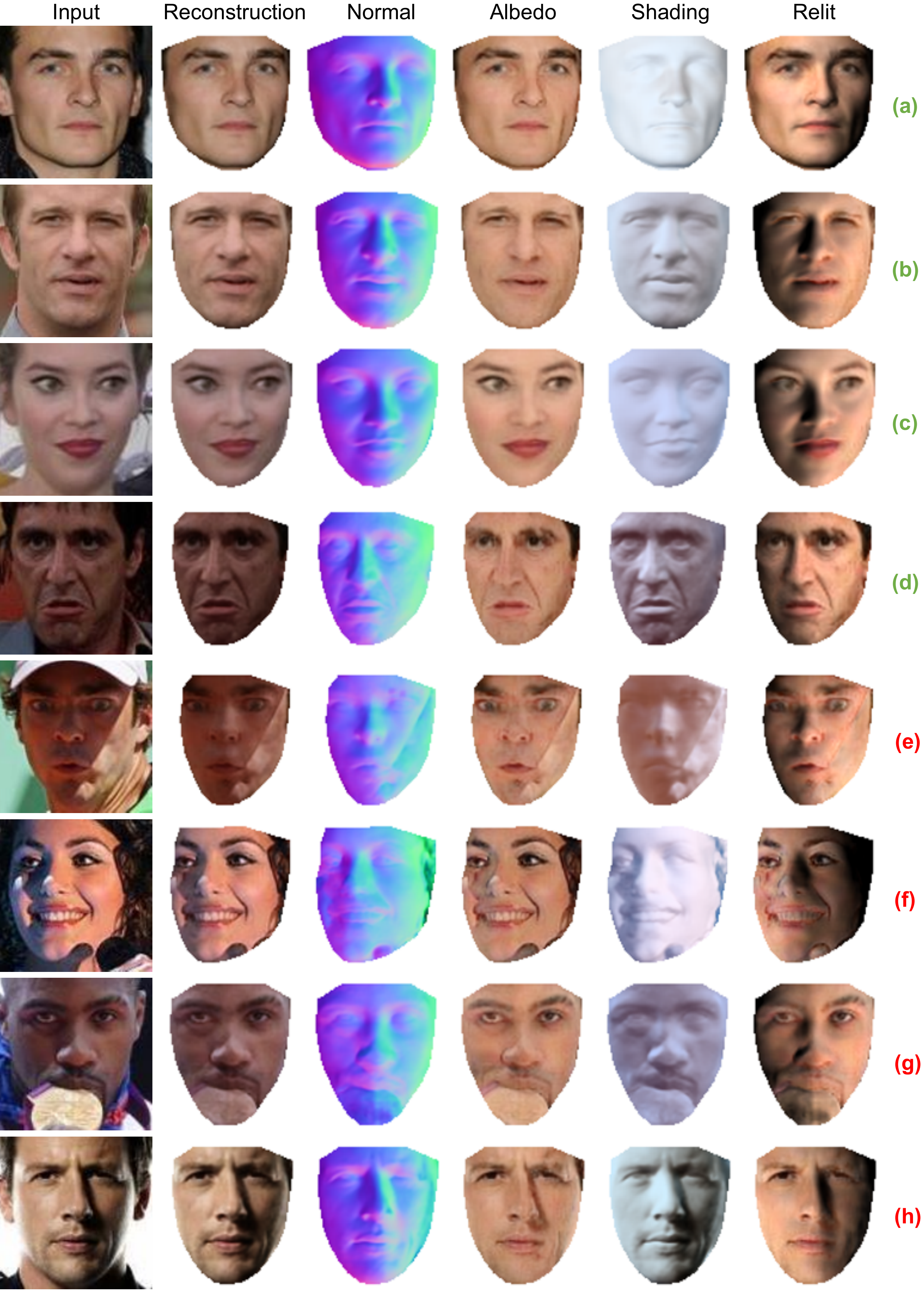}
	\caption{\small Selected results from \textbf{{\color{green!75!blue}top 5\% (a,b,c,d)}} and \textbf{\textcolor{red}{worst 5\% (e,f,g,h)}} reconstructed images. (Best viewed in color)}
	\vspace{-1em}
	\label{fig:my_best_worst}	
\end{figure}

\begin{figure}[]
	\centering
	\includegraphics[width=0.48\textwidth]{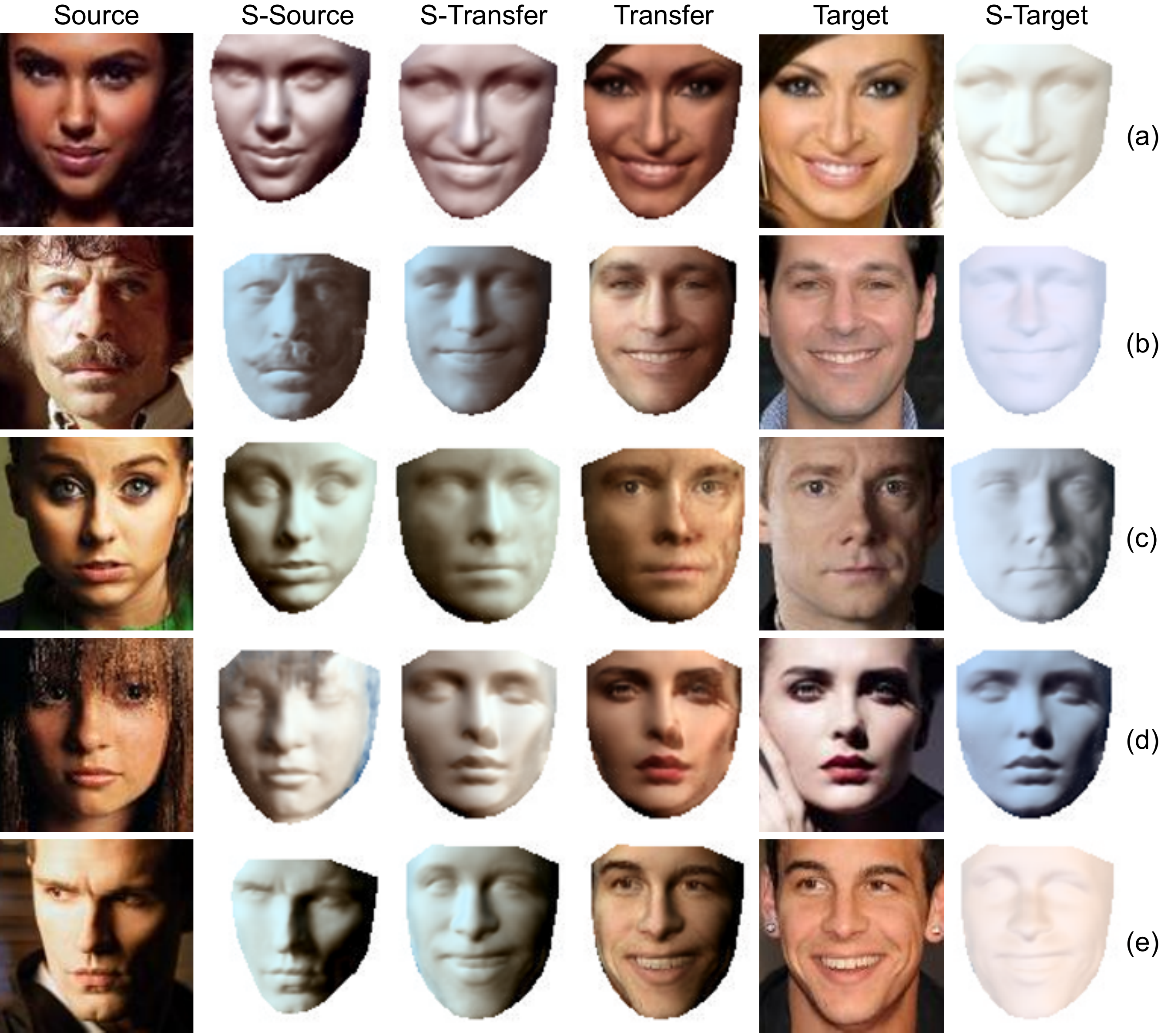}
	\caption{\small \textbf{Light transfer.} Our SfSNet allows us to transfer lighting of the `Source' image to the `Target' image to produce `Transfer' image. `S' refers to shading. (Best viewed in color)}
	\vspace{-0.5em}
	\label{fig:light_tran}	
\end{figure}

\subsection{Evaluation of Light Estimation}
\label{sec:light_hao}
\begin{table}[!h]
	\centering
	\begin{adjustbox}{max width=0.48\textwidth}
		\begin{tabular}{|c|c|c|c|}
			\hline
			Algorithm & top-1\%             & top-2\%             & top-3\%             \\ \hline
			SIRFS log \cite{barron2015shape} & 60.72 & 79.65 & 87.27 \\ \hline
			LDAN \cite{zhou2017label}     & 65.87       & 85.17     & 92.46          \\ \hline
			SfSNet    & \textbf{78.44} & \textbf{89.44} & \textbf{92.64} \\ \hline
		\end{tabular}
	\end{adjustbox}
	\caption{\textbf{Light Classification Accuracy on MultiPIE dataset.} SfSNet significantly outperforms `LDAN'.}
	\vspace{-1em}
	\label{tab:light_hao}
\end{table}

We evaluate the quality of the estimated lighting using MultiPIE dataset \cite{gross2010multi} where each of the 250 individuals is photographed under 19 different lighting conditions. We perform 19-way classification, to check the consistency of the estimated lighting as described in \cite{zhou2017label} and compare with their proposed algorithm `LDAN'. `LDAN' estimates lighting independently from a single face image using adversarial learning. Results in Table \ref{tab:light_hao} shows that we improve top-1\% classification accuracy by 12.6\% over `LDAN'.

\section{Results on CelebA}
\label{sec:sfs_res}
In Figure \ref{fig:my_best_worst} we provide sample results on CelebA test data from the best 5\% and worst 5\% reconstructed images respectively. For every test face, we also relight the face using a directional light source that highlights the flaws in the decomposition. As expected the best results are for frontal faces with little or no expression and easy ambient lighting as shown in Figure \ref{fig:my_best_worst} (a-d). The worst reconstructed images have large amounts of cast shadows, specularity and occlusions as shown in Figure \ref{fig:my_best_worst} (e-h). However, the recovered normal and lighting are still reasonable. We also show interesting results on light transfer in Figure \ref{fig:light_tran}, which also highlights the quality of the decomposition. Note that the examples shown in (c) and (d) are particularly hard as source and target images have opposite lighting directions. More qualitative results on CelebA and comparison with \cite{adobe,mofa,sela2017unrestricted} are provided in the Appendix \ref{sec:extra}.

\section{Ablation Studies}
We analyze the relative importance of mixed data training with `SfS-supervision' compared to learning from synthetic data alone. We also contrast the SfSNet architecture with skip-connection based networks. For ablation studies, we consider photometric reconstruction loss (Recon. Error) and lighting classification accuracy (Lighting Acc.) as performance measures.\\

\noindent
\textbf{Role of `SfS-supervision' training:}
\label{sec:mix_res}
To analyze the importance of our mixed data training we consider the SfSNet architecture and compare its performance using different training paradigms. We consider the following:\\
\textbf{SfSNet-syn:}  We train SfSNet on synthetic data only.\\
\textbf{SkipNet-syn:} We observe that our residual block based network can not generalize well on unseen real world data when trained on synthetic data, as there is no direct skip connections that can transfer high frequencies from input to output. However a skip connection based encoder-decoder network can generalize on unseen real world data. Thus we consider a skip connection based network, `SkipNet', which is similar in structure with the network presented in \cite{adobe}, but with increased capacity and skip connections. We train `SkipNet' on synthetic data only and this training paradigm is similar to \cite{shi2016learning}, which also uses a skip-connection based network for decomposition in ShapeNet objects.\\
\textbf{SfSNet:} We use our `SfS-supervision' to train our SfSNet, where `pseudo-supervision' is generated by `SkipNet'.
\begin{table}[!h]
	\centering
	\begin{adjustbox}{max width=0.48\textwidth}
		\begin{tabular}{|c|c|c|c|c|c|}
			\hline
			\multirow{2}{*}{Training  Paradigm} & \multicolumn{2}{l|}{Recon. Error} & \multicolumn{3}{l|}{Lighting Acc.}  \\ \cline{2-6} 
			& MAE                 & RMSE                & Rank 1           & Rank 2           & Rank 3           \\ \hline
			SkipNet-syn                  & 42.83               & 48.22               & 54.86\%          & 76.78\%          & 85.76\%          \\ \hline
			SfSNet-syn                    & 48.54               & 58.13               & 63.88\%          & 80.52\%          & 87.24\%          \\ \hline
			SfSNet                              & \textbf{10.99}      & \textbf{13.55}      & \textbf{78.44\%} & \textbf{89.52\%} & \textbf{92.64\%} \\ \hline
		\end{tabular}
	\end{adjustbox}
	\caption{\textbf{Role of `SfS-supervision' training.} `SfS-supervision' outperforms training on synthetic data only.}
	\vspace{-0.7em}
	\label{tab:ab_train}
\end{table}

Note that another alternative is training on synthetic data and fine-tuning on real data. It has been shown in \cite{adobe} that it is not possible to train the network on real data alone by using only reconstruction loss, as the ambiguities in the decomposition can not be constrained, leading to a trivial solution. We also find that the same argument is true in our experiments. Thus we compare our `SfS-supervision' training paradigm with only synthetic data training in Table \ref{tab:ab_train}. The results show that our `SfS-supervision' improves significantly over the `pseudo-supervision' used from SkipNet, indicating that we are successfully using shading information to add details in the reconstruction.\\

\noindent
\textbf{Role of SfSNet architecture:}
\label{sec:resVsskip}
We evaluate the effectiveness of our proposed architecture against a skip connection based architecture. Our proposed architecture estimates lighting from image, normal and albedo, as opposed to a skip connection based network which estimates lighting directly from the image only. SkipNet described in the Appendix based on \cite{adobe} does not produce a good decomposition because of the fully connected bottleneck. Thus we compare with a fully convolutional architecture with skip connection, similar to Pix2Pix  \cite{isola2016image}, which we refer to as Skipnet+. This network has one encoder, two decoders for normal and albedo and a fully connected layer from the output of the encoder to predict light (see Appendix \ref{sec:arch_skip+} for details).


In Table \ref{tab:ab_skip2} we show that our SfSNet outperforms SkipNet+, also trained using the `SfS-supervision' paradigm. Although reconstruction error is similar for both networks, SfSNet predicts better lighting than `SkipNet+'. This improved performance can be attributed to the fact that SfSNet learns an informative latent subspace for albedo and normal, which is further utilized along with image features to estimate lighting. Whereas in the case of the skip connection based network, the latent space is not informative as high frequency information is directly propagated from input to output bypassing the latent space. Thus lighting parameters estimated only from the latent space of the image encoder fail to capture the illumination variations.

\begin{table}[!h]
	\centering
	\begin{adjustbox}{max width=0.48\textwidth}
		\begin{tabular}{|c|c|c|c|c|c|}
			\hline
			\multirow{2}{*}{Training  Paradigm} & \multicolumn{2}{l|}{Recon. Error} & \multicolumn{3}{l|}{Lighting Acc.}      \\ \cline{2-6} 
			& MAE             & RMSE            & top-1\%           & top-2\%           & top-3\%           \\ \hline
			SkipNet+                     & 11.33           & 14.42           & 67.70\%          & 85.08\%          & 90.34\%          \\ \hline
			SfSNet                              & \textbf{10.99}  & \textbf{13.55}  & \textbf{78.44\%} & \textbf{89.52\%} & \textbf{92.64\%} \\ \hline
		\end{tabular}
	\end{adjustbox}	
	\caption{\textbf{SfSNet vs SkipNet+}. Proposed SfSNet outperforms a skip connection based SkipNet+ which estimates lighting directly from the image.}
	\vspace{-1.5em}
	\label{tab:ab_skip2}
\end{table}

\section{Conclusion}
\label{sec:discussion}
In this paper we introduce a novel architecture SfSNet, which learns from a
mixture of labeled synthetic and unlabeled real images to solve the problem of inverse
face rendering. SfSNet is inspired by a physical rendering model and utilizes
residual blocks to disentangle normal and albedo into separate subspaces. They
are further combined with image features to estimate lighting. Detailed
qualitative and quantitative evaluations show that SfSNet significantly
outperforms state-of-the-art methods that perform inverse rendering and methods
that only estimate the normal or lighting.
\\

\noindent \textbf{Acknowledgment}
This research is supported by the National Science Foundation under grant no. IIS-1526234. We thank Hao Zhou and Rajeev Ranjan for helpful discussions, Ayush Tewari for providing visual results of MoFA, and Zhixin Shu for providing test images of Neural Face.

\small
\bibliographystyle{ieee}
\bibliography{sfsB}

\newpage 
\section{Appendix}
\label{sec:supp}


\subsection{SfSNet Architecture}
\label{sec:arch_sfs}
\begin{figure}[!h]
	\centering
	\includegraphics[width=0.48\textwidth]{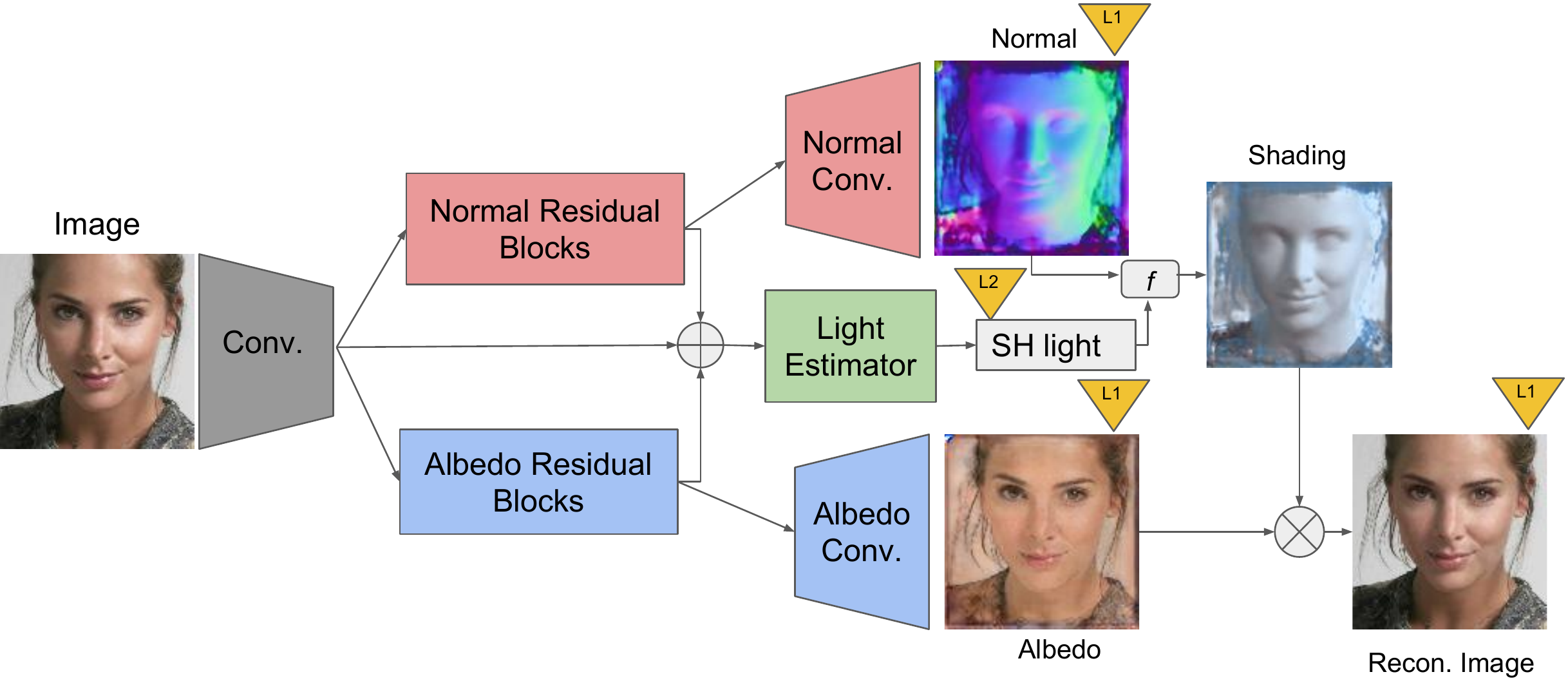}
	\caption{\small \textbf{SfSNet Architecture}.}
	\vspace{-1em}
	\label{fig:SfSNet}	
\end{figure}

The schematic diagram of our SfSNet is again shown in Figure \ref{fig:SfSNet} for reference. Our input, normal and albedo is of size $128 \times 128$. Below we provide the details of each of the blocks of SfSNet.\\

\noindent
\textbf{`Conv.'}: C64(k7) - C128(k3) - C*128(k3)\\
`CN(kS)' denotes convolution layers with N $S \times S$ filters with stride 1, followed by Batch Normalization and ReLU. `C*N(kS)' denotes only convolution layers with N $S \times S$ filters with stride 2, without batch Normalization. The output of `Conv' layer produces a blob of spatial resolution $128 \times 64 \times 64$.\\

\noindent
\textbf{`Normal Residual Blocks'}: 5 ResBLK - BN - ReLU\\
This consists of 5 Residual Blocks, `BesBLK's, all of which operate at a spatial resolution of $128 \times 64 \times 64$, followed by Batch Normalization (BN) and ReLU. Each `ResBLK' consists of BN - ReLU - C128 - BN - ReLU - C128.\\


\noindent
\textbf{`Albedo Residual Blocks'}: Same as `Normal Residual Blocks' (weights are not shared).\\

\noindent
\textbf{`Normal Conv'.}: BU -  CD128(k1)  - C64(k3) - C*3(k1)\\
`BU' refers to Bilinear up-sampling that converts $128 \times 64 \times 64$ to $128 \times 128 \times 128$.`CN(kS)' represents convolution layers with N $S \times S$ filters with stride 1, followed by Batch Normalization and ReLU. `C*N(kS)' represents only convolution layer with N $S \times S$ filters with stride 1. The network produces a normal map as output.\\

\noindent
\textbf{`Albedo Conv.'}: Same as `Normal Conv.' (weights are not shared).\\

\noindent
\textbf{`Light Estimator'}: It first concatenates the responses of `Conv', `Normal Residual Blocks' and `Albedo Residual Blocks' to produce a blob of spatial resolution $384 \times 64 \times 64$. This is further processed by 128 $1 \times 1$ convolutions, Batch Normalization, ReLU, followed by Average Pooling over $64 \times 64$ spatial resolution to produce 128 dimensional features. This 128 dimensional feature is passed through a fully connected layer to produce 27 dimensional spherical harmonics coefficients of lighting. Our model and code is available for research purposes at {\small \url{https://senguptaumd.github.io/SfSNet/}}.

\subsection{SkipNet Architecture}
\label{sec:arch_skip}
\begin{figure}[t]
	\centering
	\includegraphics[width=0.48\textwidth]{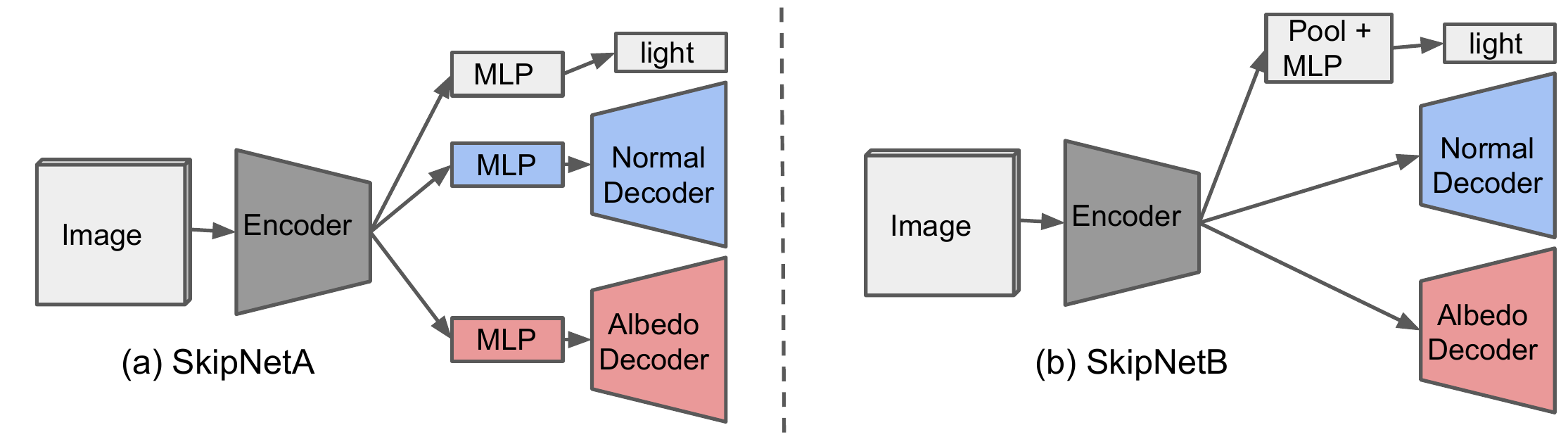}
	\caption{\small \textbf{SkipNet and SkipNet+ Network Architectures}.}
	\vspace{-1em}
	\label{fig:skipnet}	
\end{figure}

The schematic diagram of SkipNet is shown in Figure \ref{fig:skipnet}(a). SkipNet is based on the network used in \cite{adobe} with more capacity and skip connections.  Similar to SfSNet the input is $128 \times 128$; `Normal Decoder' and `Albedo Decoder' produces normal and albedo maps. Normal, albedo and `light' is also used to produce shading and the reconstructed image similar to Figure \ref{fig:SfSNet}. Since that part of the architecture does not contain any trainable parameters we omit them in the figure for clarity. Note that the skip connections between encoder and decoder exist, which is also not shown in the figure. Details of SkipNet are provided below:\\

\noindent
\textbf{Encoder}: C*64(k4) - C128(k4) - C256(k4) - C256(k4) - C256(k4) - fc256\\
`CN(kS)' represents convolution layers with N $S \times S$ filters with stride 2, followed by Batch Normalization and ReLU. `C*N(kS)' is`CN(ks)' without Batch Normalization. All ReLUs are leaky with slope 0.2. 'fc256' is a fully connected layer that produces a 256 dimensional feature.

\noindent
\textbf{MLP}: Contains a fully connected layer to take the response of Encoder and separate it into 256 dimensional features for `Normal Decoder', `Albedo Decoder' and `light'. For `Normal Decoder' and `Albedo Decoder' a 256 dimensional feature is further up-sampled to form a blob of shape $256 \times 4 \times 4$. For 'light' the 256 dimensional feature is passed through a fully connected network to produce 27 dimensional spherical harmonics coefficients.

\noindent
\textbf{Decoder (Normal and Albedo)}: CD256(k4) - CD256(k4) - CD256(k4) - CD128(k4) - CD64(k4) - C*3(k1)
Both `Normal Decoder' and `Albedo Decoder' consists of the same architecture without weight sharing. `CDN(kS)' represents a de-convolution layer with N $S \times S$ filters operated with stride 2, followed by Batch Normalization and ReLU. `C*3(k1)' consists of 3 $1 \times 1$ convolution filters with stride 1 to produce Normal or Albedo. Skip connections are present between encoders and decoders similar to \cite{isola2016image,sela2017unrestricted}.

\subsection{SkipNet+}
\label{sec:arch_skip+}
SkipNet+ is very similar to SkipNet, but with larger capacity and without a fully connected bottleneck `MLP' as shown in Figure \ref{fig:skipnet}(b). The Details of the network are shown below.

\noindent
\textbf{Encoder}: Co64(k3) - Co64(k1) - C64(k3) - Co64(k1) - C128(k3) - Co128(k1) - C256(k3) - Co256(k1) - C256(k3) - Co256(k1) - C256(k3) \\
`CN(kS)' represents a convolution layer with N $S \times S$ filters with stride 2, followed by Batch Normalization and ReLU. `CoN(kS)' is similar to `CN(kS)' but with stride 1. All ReLUs are leaky with slope 0.3. The output of the Encoder is a feature of spatial resolution $256 \times 4 \times 4$.

\noindent
\textbf{Decoder (Normal and Albedo)}: C256(k1) - CD256(k4) - CD256(k4) - CD256(k4) - CD128(k4) - CD64(k4) - C*3(k1) \\
`CDN(kS)' represents a de-convolution layer with N $S \times S$ filters with stride 2, followed by Batch Normalization and ReLU. `CN(kS)' represents a convolution layer with N $S \times S$ filters with stride 1, followed by Batch Normalization and ReLU. `C*3(k1)' consists of 3 $1 \times 1$ convolution filters to produce Normal or Albedo. Skip-connections exists between `CN(k3)' layers of encoder and `CDN(k4)' layers of decoder.

\noindent
\textbf{light}: We perform Average pooling over $4 \times 4$ spatial resolution of the encoder output to produce a 256 dimensional feature. This feature is then passed through a fully connected layer to produce 27 dimensional spherical harmonics lighting.

\subsection{Spherical Harmonics}
\label{sec:sph}
In this section, we define the image generation process under lambertian reflectance following equation \eqref{eq:ren1}. Let the normal be $n(p)=[x,y,z]^T$ at pixel $p$.  Then the 9 dimensional spherical harmonics basis $Y(p)$ at pixel $p$ is expressed as:
\begin{equation}
Y = [Y_{00}, Y_{10}, Y^{e}_{11} Y^{0}_{11}, Y_{20}, Y^{e}_{21}, Y^{o}_{21}, Y^{e}_{22}, Y^{o}_{22}]^T,
\end{equation}
where
\begin{eqnarray}
Y_{00} &= \frac{1}{\sqrt{4\pi}} \qquad  \qquad \qquad \qquad Y_{10} &=\sqrt{\frac{3}{4\pi}}z \nonumber \\
Y_{11}^{e} &= \sqrt{\frac{3}{4\pi}}x \qquad \qquad \qquad Y_{11}^{o} &=\sqrt{\frac{3}{4\pi}}y \nonumber \\
Y_{20} &= \frac{1}{2}\sqrt{\frac{5}{4\pi}}(3z^2-1) \qquad   Y_{21}^{e} &= 3\sqrt{\frac{5}{12\pi}}xz  \\
Y_{21}^{o} &= 3\sqrt{\frac{5}{12\pi}}yz   \qquad \qquad Y_{22}^{e}&= \frac{3}{2}\sqrt{\frac{5}{12\pi}}(x^2-y^2) \nonumber \\   
Y_{22}^{0} &= 3\sqrt{\frac{5}{12\pi}}xy \qquad \qquad  \nonumber
\end{eqnarray}
Then the intensity at pixel $p$ is defined as:
\begin{equation}
I(p) = f_{render}(A(p),N(p),L) = A(p) (Y(p)^{T} L),
\end{equation}
where $A(p)$ is the albedo at pixel $p$, and $L$ is the lighting parameter denoting coefficients of spherical harmonics basis. Note that, the above equations are only for one of the RGB channels and can be repeated independently for 3 channels.

Next we define the reconstruction loss. Let $I(p)$ be the original image intensity and $\tilde{N}(p)$, $\tilde{A}(p)$ be the inferred normal and albedo by SfSNet at pixel $p$. Let $\tilde{L}$ be the 27 dimensional spherical harmonic coefficients also inferred by SfSNet. The reconstruction loss is defined as:
\begin{eqnarray}
E_{recon} = \sum_{p} | I(p) - f_{render}(\tilde{A}(p),\tilde{N}(p),\tilde{L})|.
\end{eqnarray}

\subsection{More Qualitative Comparisons} 
\label{sec:extra}
\begin{figure*}[]
	\centering
	\includegraphics[width=0.98\textwidth]{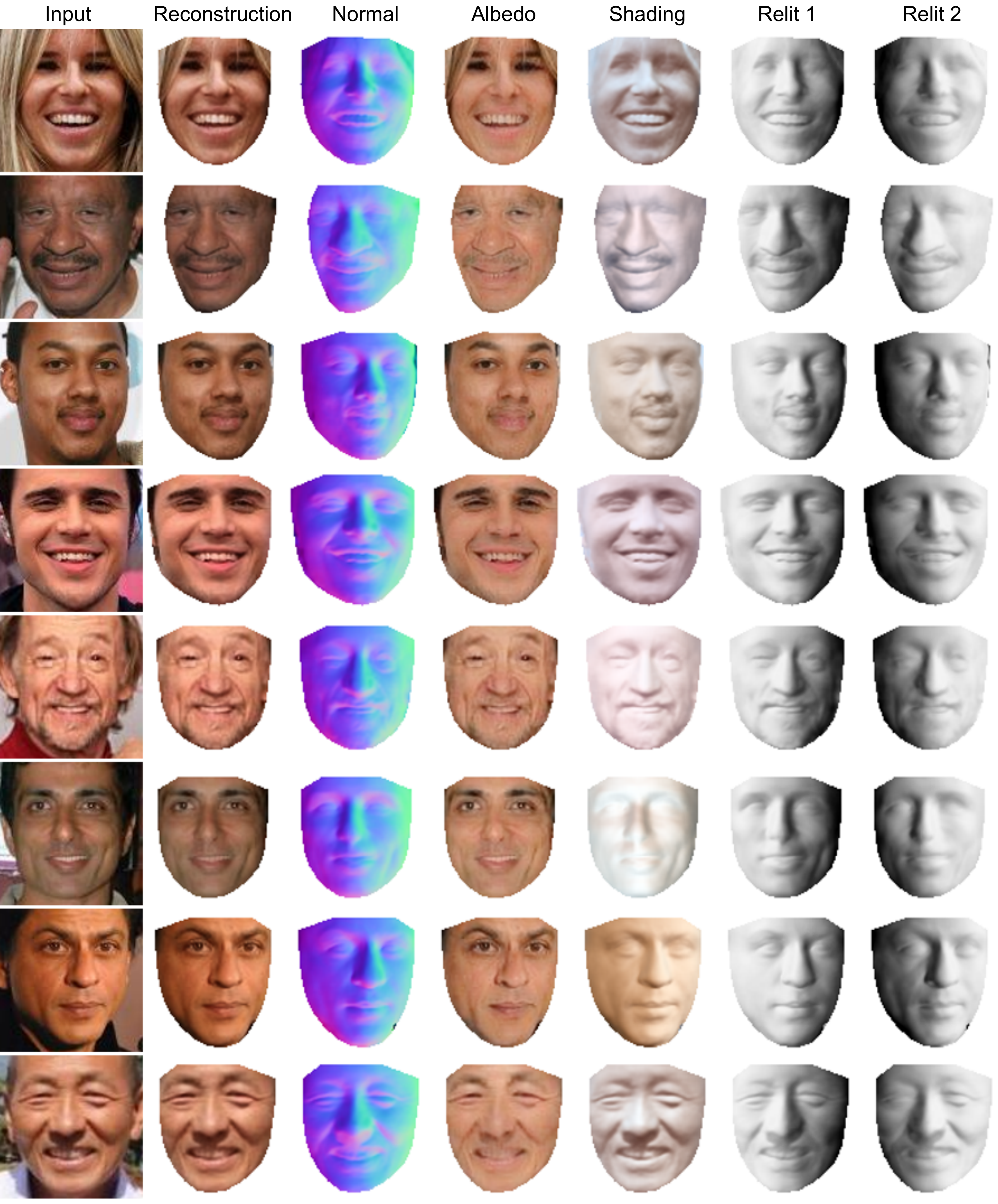}
	\caption{\small Results of SfSNet on CelebA. `Relit' images are generated by directional lighting and uniform albedo to highlight the quality of the reconstructed normals. (Best viewed in color)}
	\label{celeba1}
	\vspace{-1em}
\end{figure*}

\begin{figure*}[]
	\centering
	\includegraphics[width=0.98\textwidth]{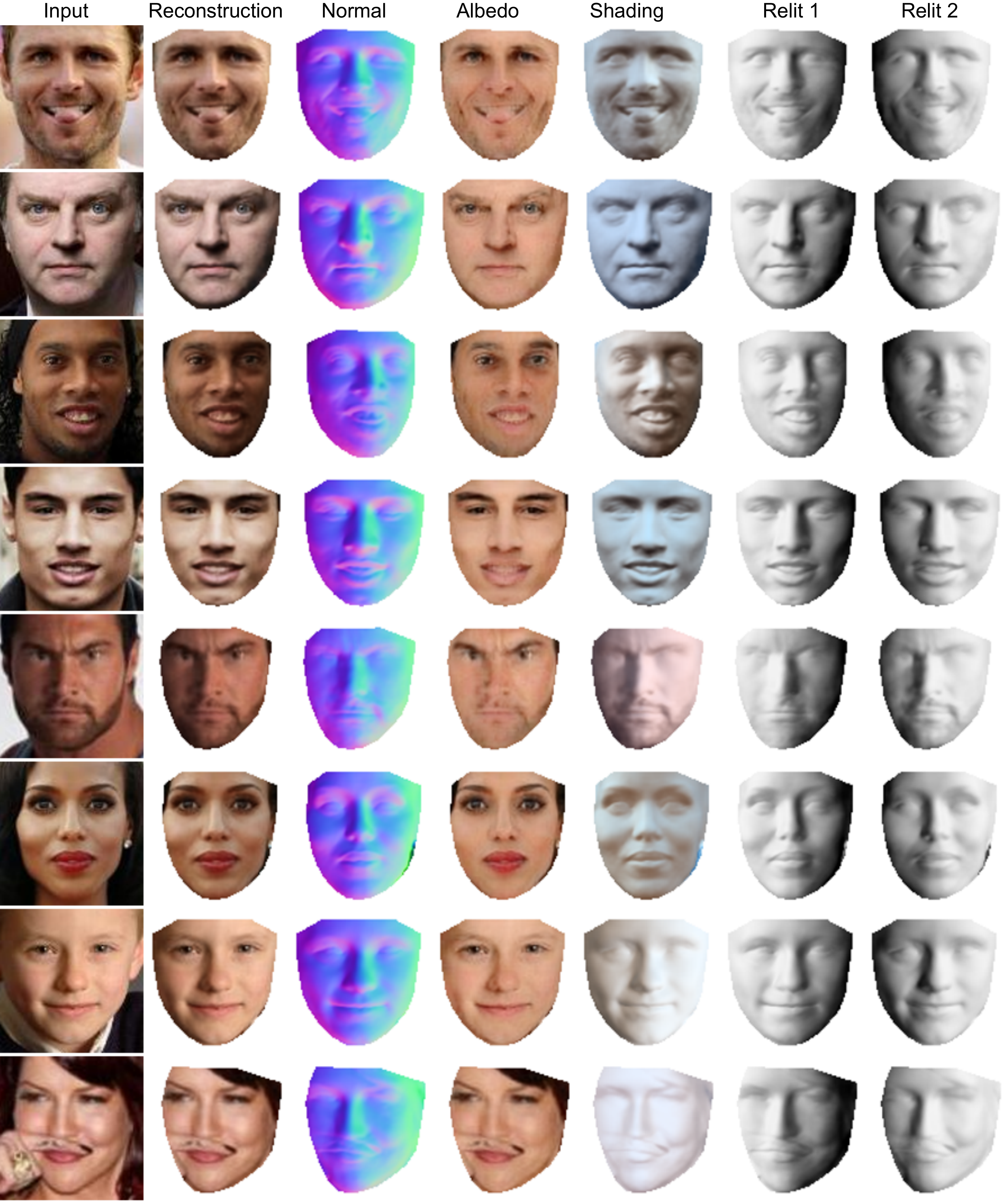}
	\caption{\small Results of SfSNet on CelebA. `Relit' images are generated by directional lighting and uniform albedo to highlight the quality of the reconstructed normals. (Best viewed in color)}
	\label{celeba2}
	\vspace{-1em}
\end{figure*}

\noindent \textbf{SfSNet on CelebA:} In Figures \ref{celeba1} and \ref{celeba2} we present inverse rendering results on CelebA images with our SfSNet. To visualize the quality of the reconstructed normals, we use directional lights with uniform albedo to produce `Relit' images. \\

\begin{figure*}[]
	\centering
	\includegraphics[width=0.9\textwidth]{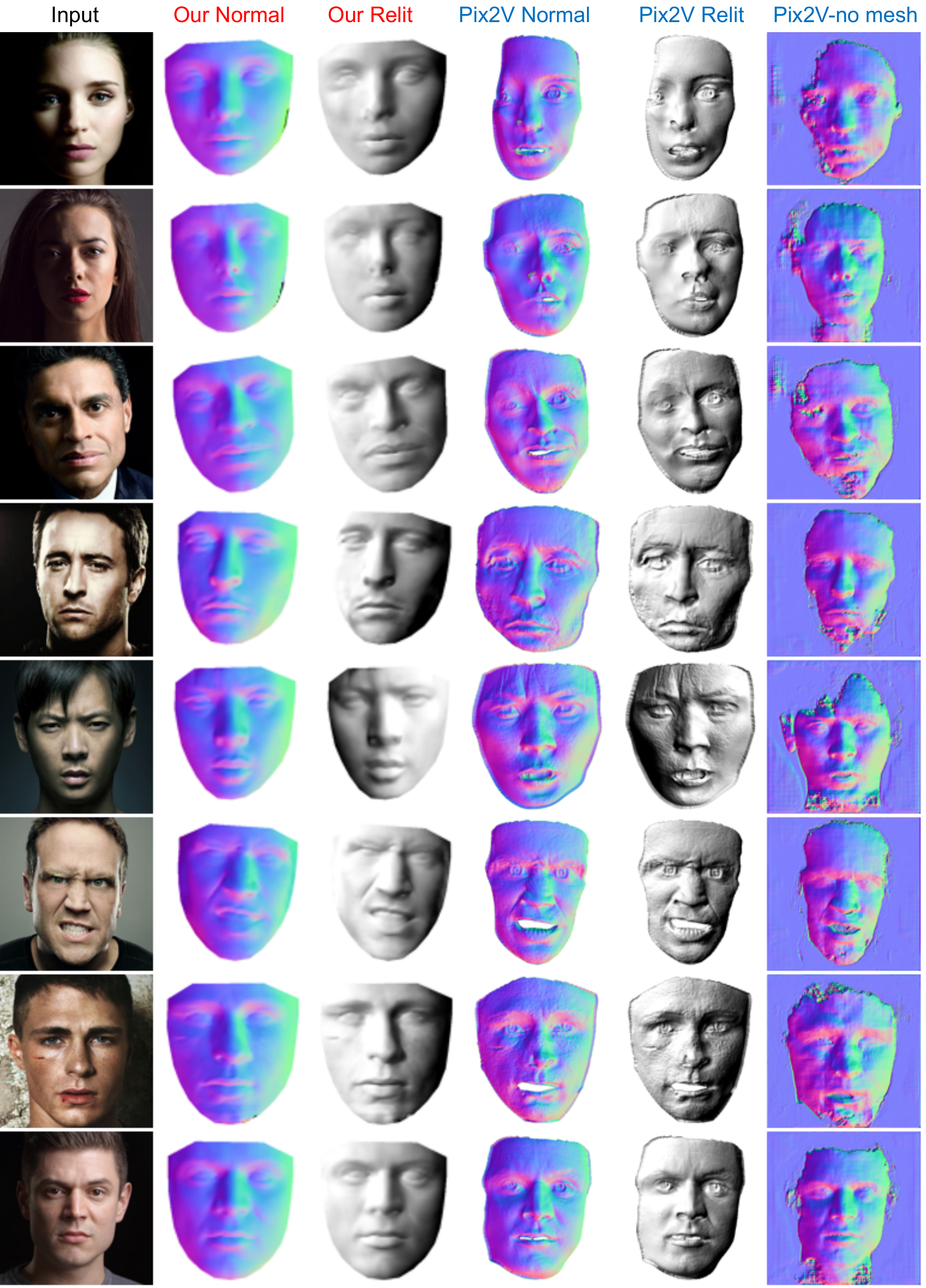}
	\caption{\small \textbf{SfSNet vs Pix2Vertex} \cite{sela2017unrestricted} on images selected by us with non-ambient illumination and expression. `Relit' images are generated by directional lighting and uniform albedo selected to highlight the quality of the reconstructed normals. (Best viewed in color)}
	\label{kimmel_hard}
	\vspace{-1em}
\end{figure*}

\begin{figure*}[]
	\centering
	\includegraphics[width=0.9\textwidth]{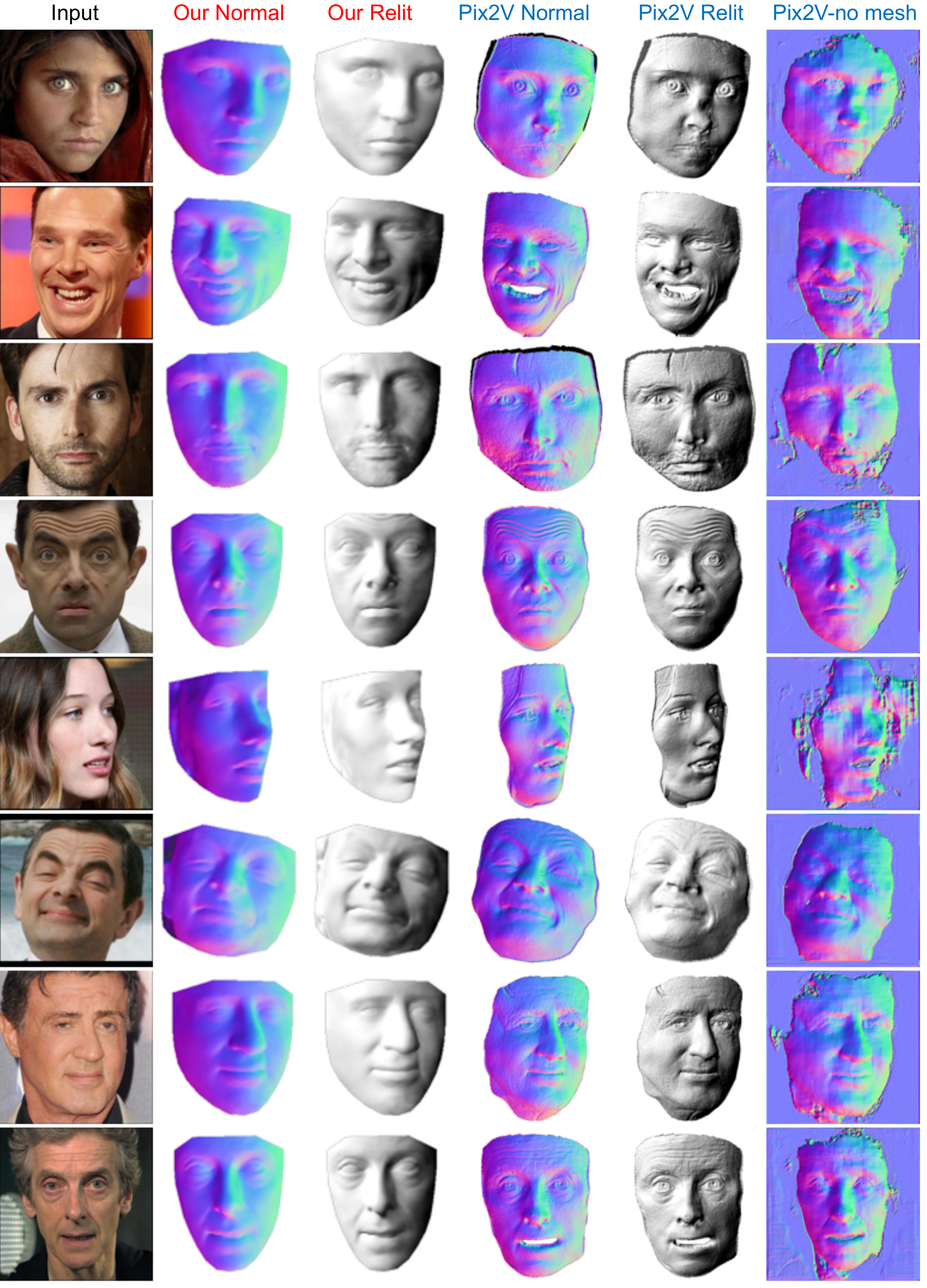}
	\caption{\small \textbf{SfSNet vs Pix2Vertex} \cite{sela2017unrestricted} on the images showcased by Sela \etal in \cite{sela2017unrestricted}. `Relit' images are generated by directional lighting and uniform albedo selected to highlight the quality of the reconstructed normals. (Best viewed in color)}
	\label{kimmel1}
	\vspace{-1em}
\end{figure*}

\begin{figure*}[]
	\centering
	\includegraphics[width=0.9\textwidth]{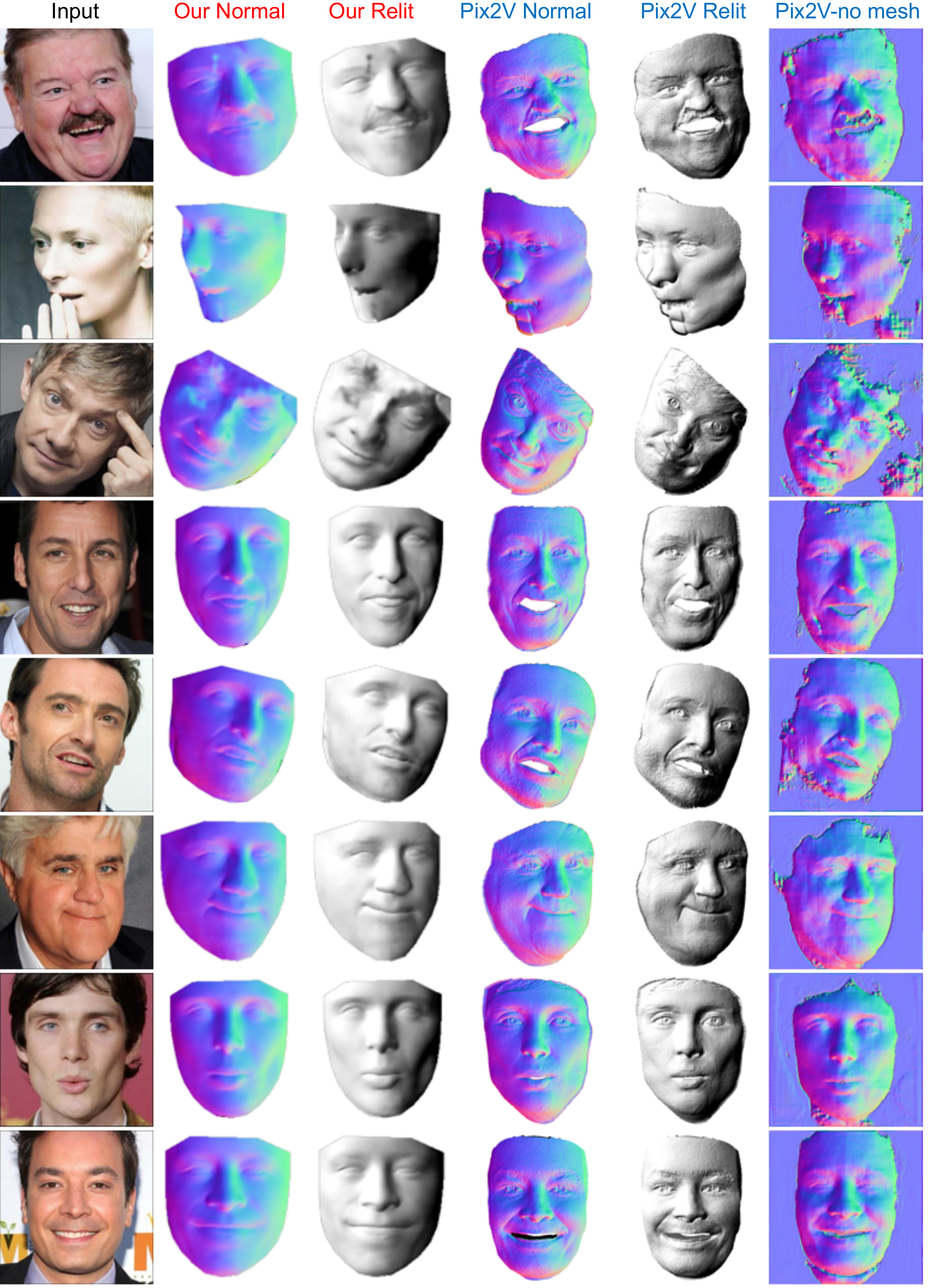}
	\caption{\small \textbf{SfSNet vs Pix2Vertex} \cite{sela2017unrestricted} on the images showcased by Sela \etal in \cite{sela2017unrestricted}. `Relit' images are generated by directional lighting and uniform albedo selected to highlight the quality of the reconstructed normals. (Best viewed in color)}
	\label{kimmel2}
	\vspace{-1em}
\end{figure*}

\begin{figure*}[]
	\centering
	\includegraphics[width=0.8\textwidth]{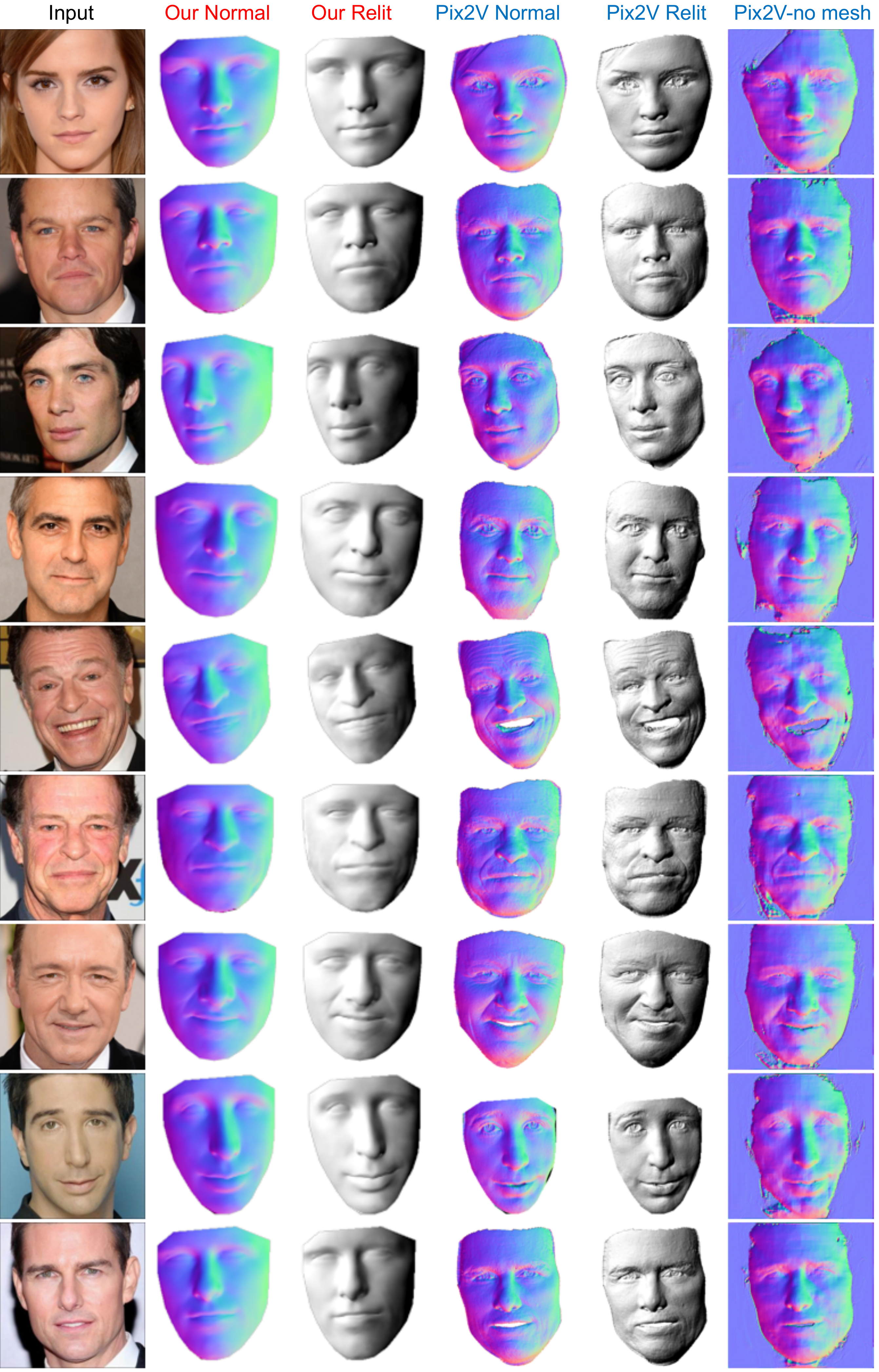}
	\caption{\small \textbf{SfSNet vs Pix2Vertex} \cite{sela2017unrestricted} on the images showcased by Sela \etal in \cite{sela2017unrestricted}. `Relit' images are generated by directional lighting and uniform albedo selected to highlight the quality of the reconstructed normals. (Best viewed in color)}
	\label{kimmel3}
	\vspace{-1em}
\end{figure*}

\noindent \textbf{SfSNet vs Pix2Vertex:} In Figure \ref{kimmel_hard} we compare SfSNet to Pix2Vertex \cite{sela2017unrestricted}. These images contain non-ambient illuminations and expressions, where surface normal recovery is much more robust for SfSNet than for Pix2Vertex. Figures \ref{kimmel1}, \ref{kimmel2} and \ref{kimmel3} also compares performance of SfSNet and Pix2Vertex on the images showcased by Sela \etal in \cite{sela2017unrestricted}. Since these images mostly contain ambient illumination, SfSNet performs comparable to Pix2Vertex. \\

\begin{figure*}[]
	\centering
	\includegraphics[width=0.55\textwidth]{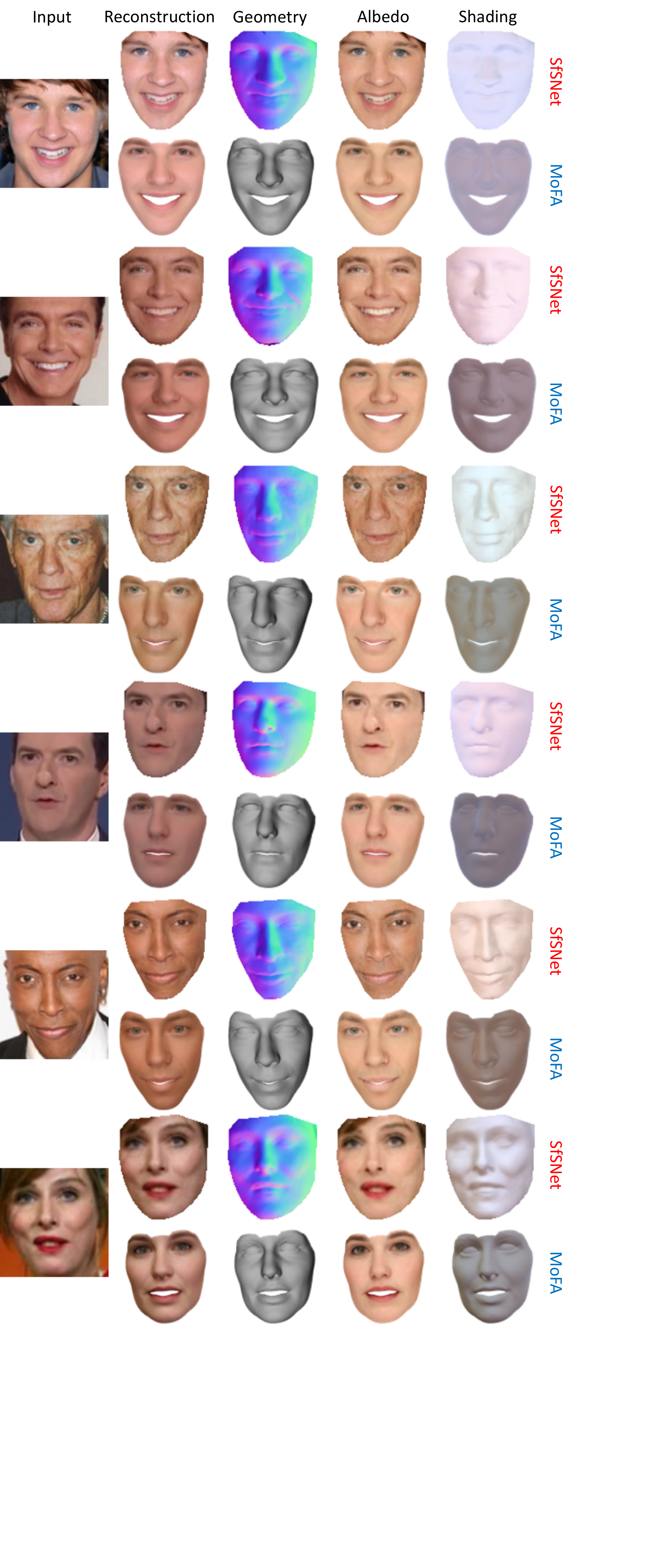}
	\caption{\textbf{Inverse Rendering.} \textbf{SfSNet vs `MoFA'} \cite{mofa} on the data provided by the authors. (Best viewed in color)}
	\vspace{-1em}
	\label{mofa1}
\end{figure*}

\begin{figure*}[]
	\centering
	\includegraphics[width=0.55\textwidth]{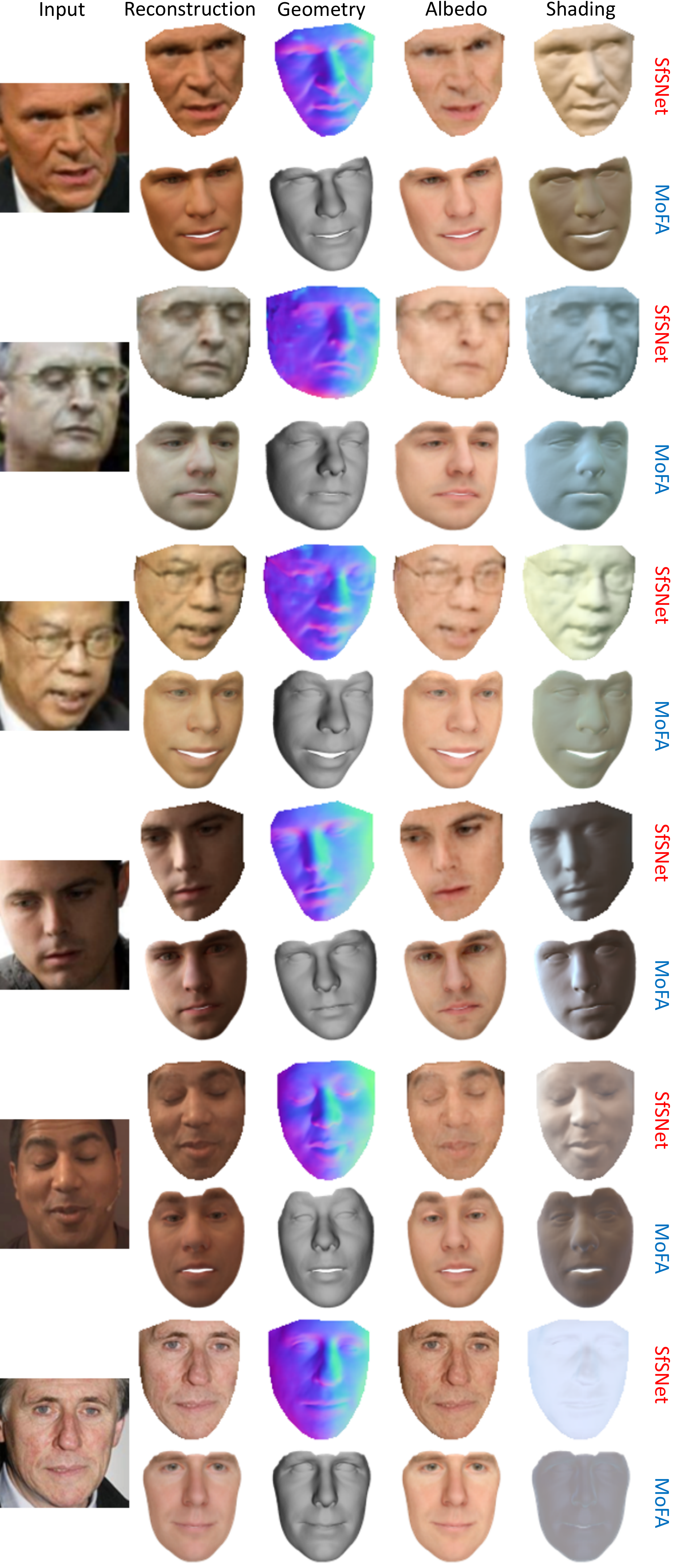}
	\caption{\textbf{Inverse Rendering.} \textbf{SfSNet vs `MoFA'} \cite{mofa} on the data provided by the authors. (Best viewed in color)}
	\vspace{-1em}
	\label{mofa2}
\end{figure*}

\begin{figure*}[]
	\centering
	\includegraphics[width=0.55\textwidth]{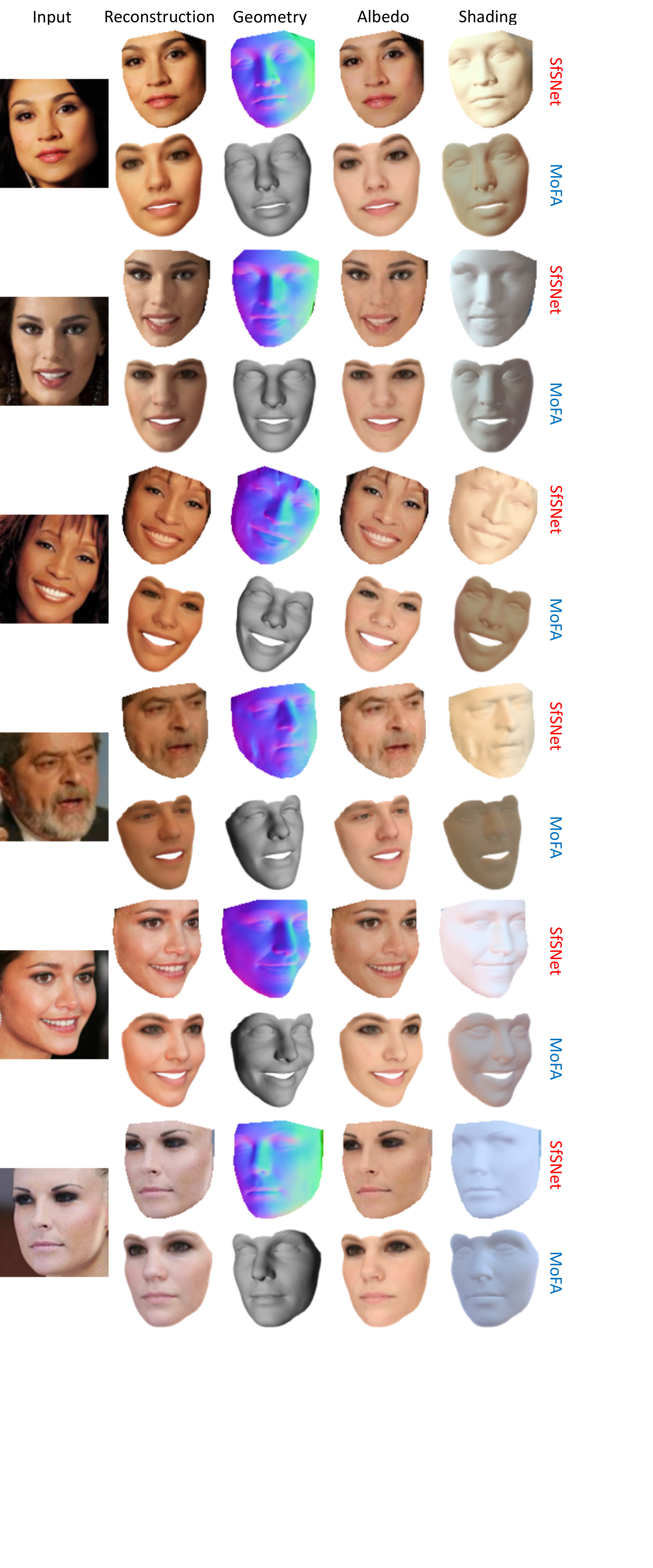}
	\caption{\textbf{Inverse Rendering.} \textbf{SfSNet vs `MoFA'} \cite{mofa} on the data provided by the authors. (Best viewed in color)}
	\vspace{-1em}
	\label{mofa3}
\end{figure*}

\noindent \textbf{SfSNet vs MoFA:} We also provide more comparison results with MoFA \cite{mofa} on the images provided by the authors in Figures \ref{mofa1}, \ref{mofa2} and \ref{mofa3}. MoFA aims to fit a 3DMM which is limited in its capability to represent real world shapes and reflectance, but can produce a full 3D mesh. Thus SfSNet reconstructs more detailed shape and reflectance than MoFA. \\

\begin{figure*}[]
	\centering
	\includegraphics[width=0.545\textwidth]{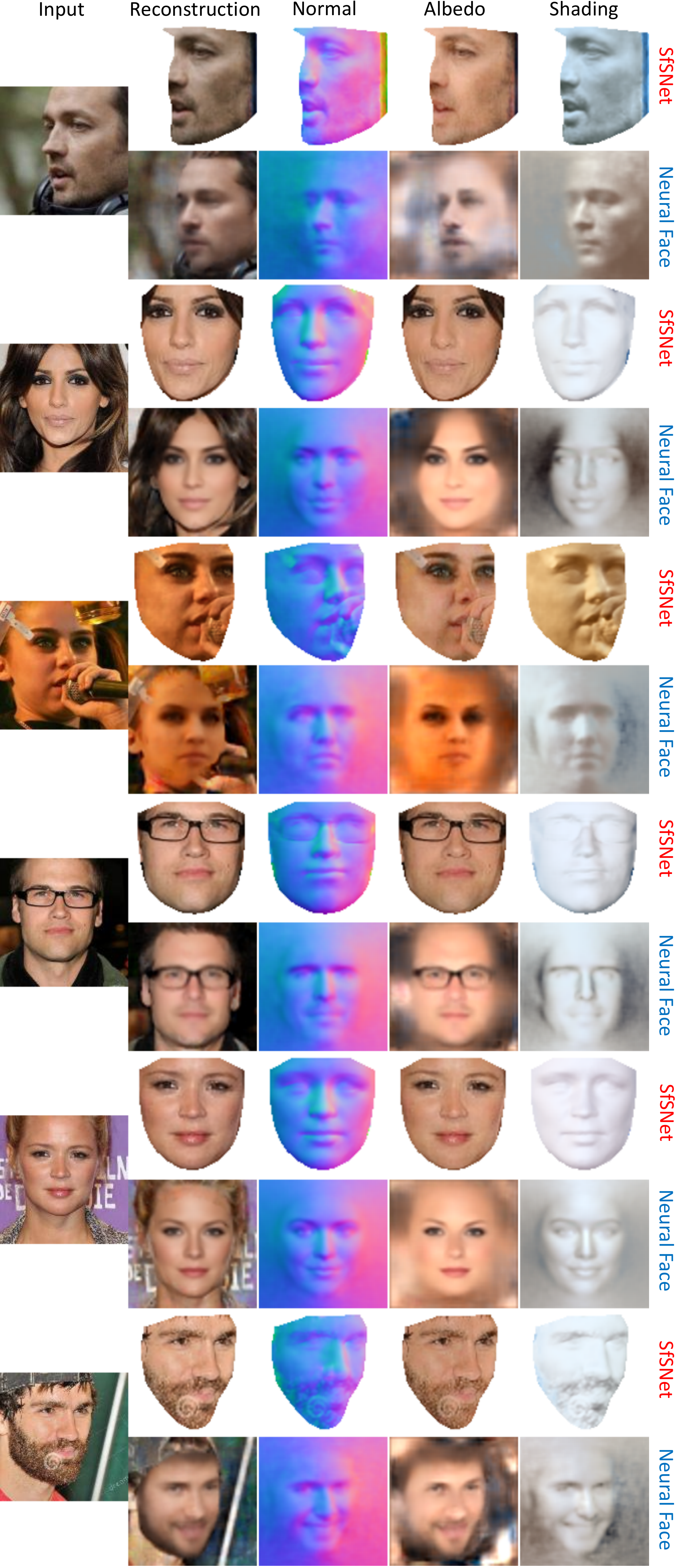}
	\caption{\small \textbf{Inverse Rendering.} \textbf{SfSNet vs `Neural Face'} \cite{adobe} on the images showcased by the authors. (Best viewed in color)}
	\label{adobe}
	\vspace{-1em}
\end{figure*}

\noindent \textbf{SfSNet vs Neural Face:} Similarly comparison with `Neural Face' \cite{adobe} in Figure \ref{adobe} on the images showcased by the authors, show that SfSNet obtains more realistic reconstruction than `Neural Face'.

\begin{figure*}[]
	\centering
	\includegraphics[width=0.9\textwidth]{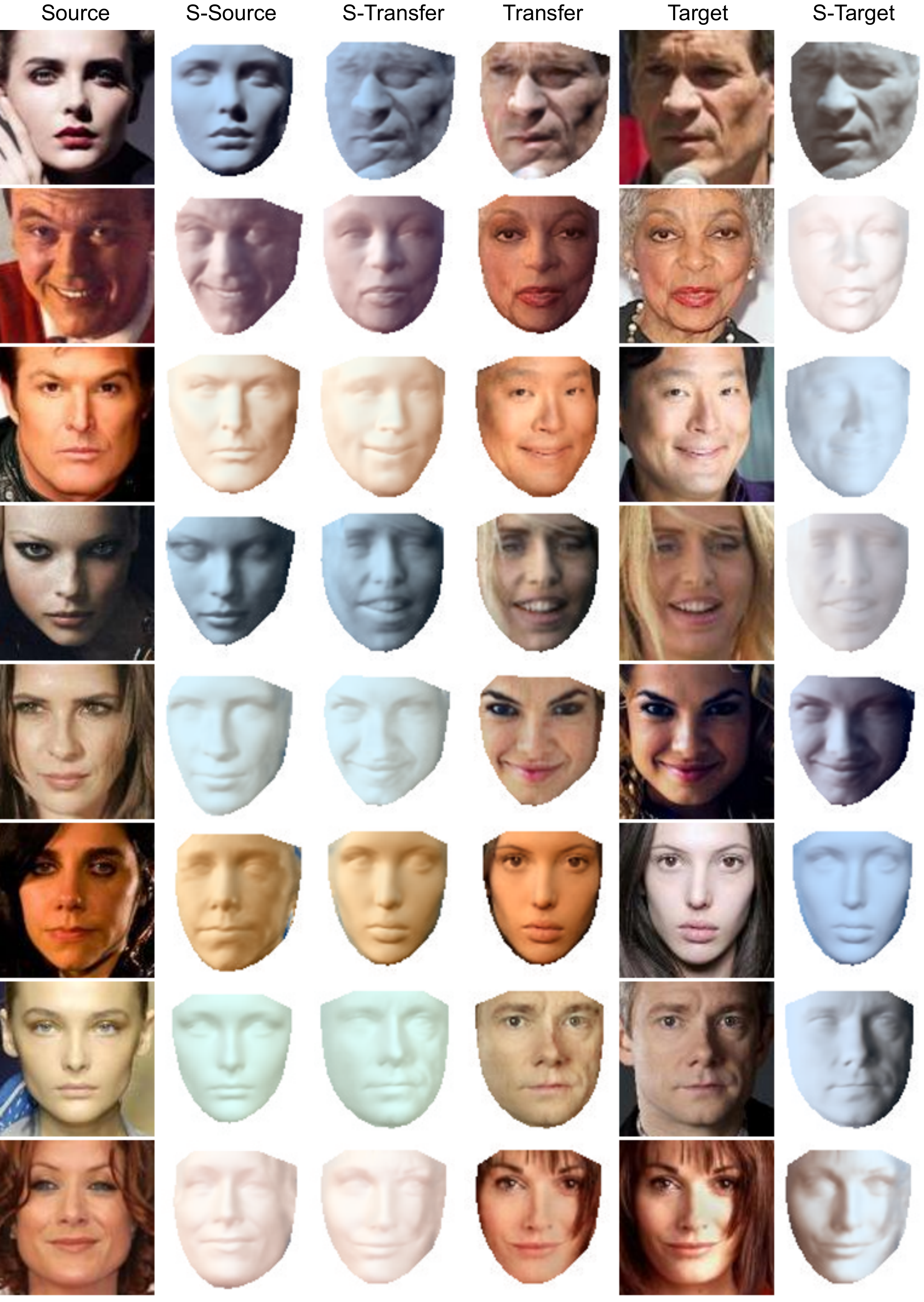}
	\caption{\small \textbf{Light transfer.} Our SfSNet allows us to transfer lighting of the `Source' image to the `Target' image to produce `Transfer' image. `S' refers to shading. (Best viewed in color)}
	\label{light_tran}
	\vspace{-1em}
\end{figure*}

\end{document}